\pdfoutput=1
\documentclass[10pt,twocolumn]{article}

\usepackage[utf8]{inputenc}
\usepackage[T1]{fontenc}
\usepackage{lmodern}
\usepackage[a4paper, margin=0.85in]{geometry}
\usepackage{graphicx}
\usepackage{longtable}
\usepackage{amsmath, amssymb}
\usepackage[hidelinks]{hyperref}
\hypersetup{pdftitle={The evolution of sex for artificial intelligence},
            pdfauthor={Giorgio F. Gilestro}}
\usepackage{microtype}
\usepackage[font=small]{caption}

\begin{document}

\twocolumn[{%
  \begin{center}
    {\LARGE\bfseries The evolution of sex for artificial intelligence}\\[0.6em]
    {\large A population-genetic framework for multigenerational model populations}\\[1.2em]
    {\normalsize Giorgio F.\ Gilestro}\\[0.3em]
    {\normalsize Department of Life Sciences, Imperial College London, London, UK}\\
    {\normalsize \href{mailto:giorgio@gilest.ro}{giorgio@gilest.ro} \,\(\cdot\)\,
     \href{mailto:g.gilestro@imperial.ac.uk}{g.gilestro@imperial.ac.uk} \,\(\cdot\)\,
     \href{https://lab.gilest.ro}{lab.gilest.ro}}\\[0.6em]
    {\normalsize \href{https://git.lab.gilest.ro/giorgio/MachineSex}{git.lab.gilest.ro/giorgio/MachineSex}}
  \end{center}
  \vspace{1.4em}
  % Reason: full-width abstract set as spaced blocks, so its three paragraphs stay distinct
  % without the indentation that the two-column body uses.
  {\setlength{\parskip}{0.7em}\setlength{\parindent}{0pt}\section*{Abstract}

Some aspects of AI development resemble a population process in which models are specialised, retrained on the output of peers, or combined by averaging weights. These practices lead to generations of models, in the biological sense studied by population genetics. Here, I develop this parallelism and interpret multigenerational model populations in terms of sexual and asexual reproduction, formally recombining the two fields. I test these analogies in an exact inheritance model, in trained networks (recurrent, feedforward and variational autoencoder generators) and in large language models, and show that they hold generally, with some measurable architecture-specific biases.

Training recursively on model output is known to lead to model collapse, a process previously described as akin to genetic drift; I develop all that follows. A minimal model of a learner retrained on its parent's output reproduces the Wright--Fisher process exactly; verified real data added to each generation play the role of immigration, with the surprising finding that the absolute number of real data samples matters, not their share, exactly as in population genetics. Training a child on the average of its parents' outputs cancels the benefit of having several parents, matching blending inheritance (and reviving Jenkin's objection to Darwin), whereas combining parents so that each keeps its strongest contribution preserves it; merged language-model specialists exceeded every parent across seeds (the Fisher--Muller effect); and lineages become reproductively isolated, losing the ability to merge at all, when they have learned conflicting conventions and not when they have merely drifted apart.

As AI societies become societies in time as well as in space, a mathematical framework for their inheritance acquires predictive power. Remarkably, that framework can be adapted almost wholesale from biology.

}
  \medskip\hrule
  \vspace{1.6em}
}]

\section*{Introduction}

Machine learning has become a population-scale phenomenon. Public repositories host millions of models (Hugging Face passed three million by 2026), most of them fine-tunes, distillations, or merges of a few foundation models, forming family trees already mapped by phylogenetic methods (1--3). \emph{Model merging}, the combination of trained parents into a new model by averaging their weights, is mainstream practice with standard tooling and thousands of hybrid checkpoints, some topping leaderboards (4--7), and its literature already speaks of ``crossover,'' ``mutation,'' and ``mate choice'' in populations of merging models that climb benchmarks (5, 8, 9) and stagnate as their members grow alike (10).

Generations are coupled through data as well as weights. Models increasingly learn from model output: populations of models are fine-tuned round after round on one another's output (11), frontier alignment pipelines are predominantly synthetic (over 98\% in documented cases; 12, 13), self-generated instruction data seed whole lineages (14), much of the public web is machine-translated (15, 16), and the stock of human text is projected to run out within a few years (17). Multi-agent systems and agent economies put many models into sustained contact (18--21). A population whose members inherit from one another, recombine, and retransmit is an evolving population in the technical sense, and the branch of biology built for that situation, the population genetics of the evolution of sex, turns out to explain what has been observed in such populations and to predict what has not (a reading anticipated by the interpretation of sex as an algorithm for mixability; 22).

Training each generation on the previous generation's output degrades it (\emph{model collapse}). Rare capabilities vanish first and the lineage drifts toward its own most common behaviour (23). That degradation follows the rule of \emph{genetic drift}, the loss of rare variants in any finite population when each generation is a finite sample of the last (the accident by which rare surnames vanish from small villages, with nothing selecting against them). The identification has been made independently for sequential inference chains before deep learning (24), for language-model text ecosystems (25), as a first-extinction law (26), and in quantitative-genetic form for self-consuming diffusion models (27). Drift is only the entry point, because population genetics is above all a theory of what keeps a finite population from decaying (immigration, recombination, selection, population structure) and of where each of those fails, and every one of them has a counterpart that whoever runs a model population (its \emph{operator}: a laboratory, a company or an automated pipeline) can switch on: real data entering each generation, merging, selection against a verifier, and the choice of which models merge with which.

Such an operator faces recurring decisions with no principled guidance. How many verified real examples does retraining need? Will combining two models compose their abilities or damage them? Can incompatibility be detected before a failed merge is paid for? When should specialists be kept separate? These are machine learning's oldest problem, \emph{continual learning} (acquiring new abilities without losing old ones; 28, 29), transposed from a single network to a population whose members inherit from one another, and each has a population-genetic answer with a number attached (how many real samples per generation, how far the average sits below the best parent, how much the parents disagree on shared inputs). Table 1 gives the correspondences the argument runs on. Fig. 1A maps the programme across three tiers (an inheritance model in simulation, trained neural networks, language models). Fig. 1B draws the change of viewpoint the framework rests on. Models are usually pictured as a society in space, contemporaries exchanging messages, but the couplings that matter here (training on model output, merging, real data entering each generation) run between generations, and a society coupled in time is what population genetics describes. As a growing share of what each model learns comes from earlier models, that is the society AI is becoming.

\begin{figure*}[p]\centering  % fig1
\includegraphics[width=\textwidth,height=0.28\textheight,keepaspectratio]{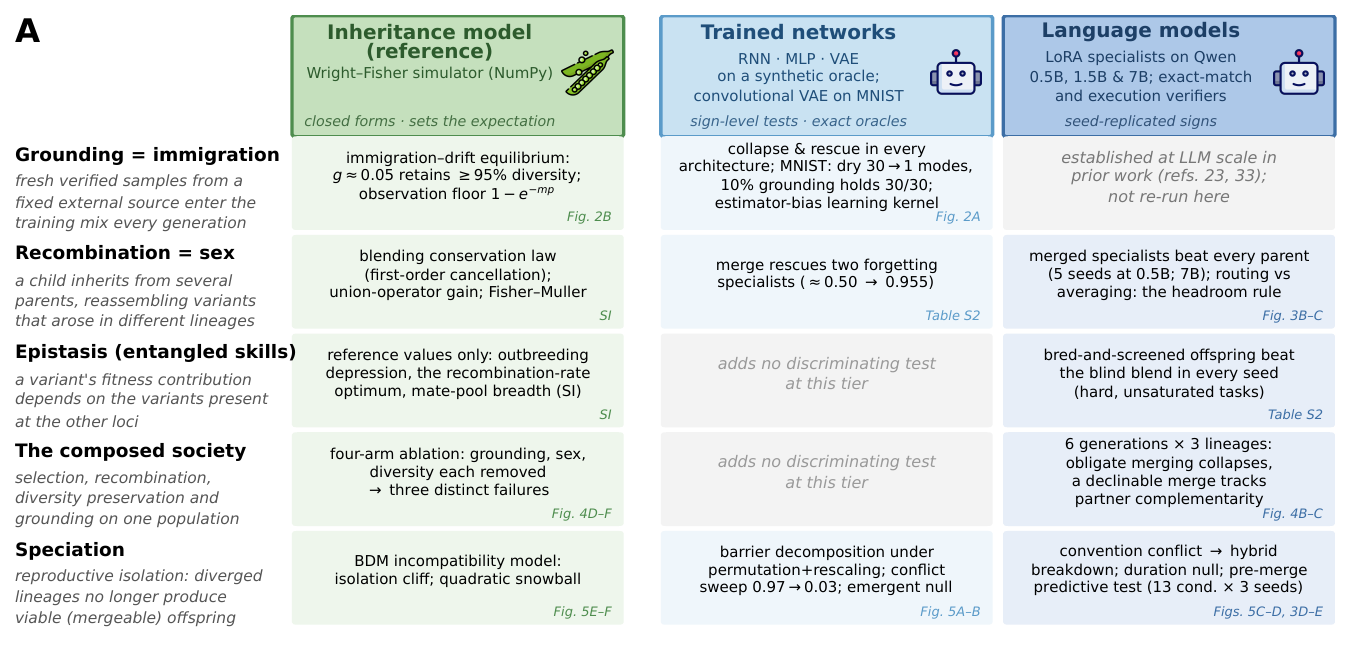}\\[6pt]
\includegraphics[width=\textwidth,height=0.28\textheight,keepaspectratio]{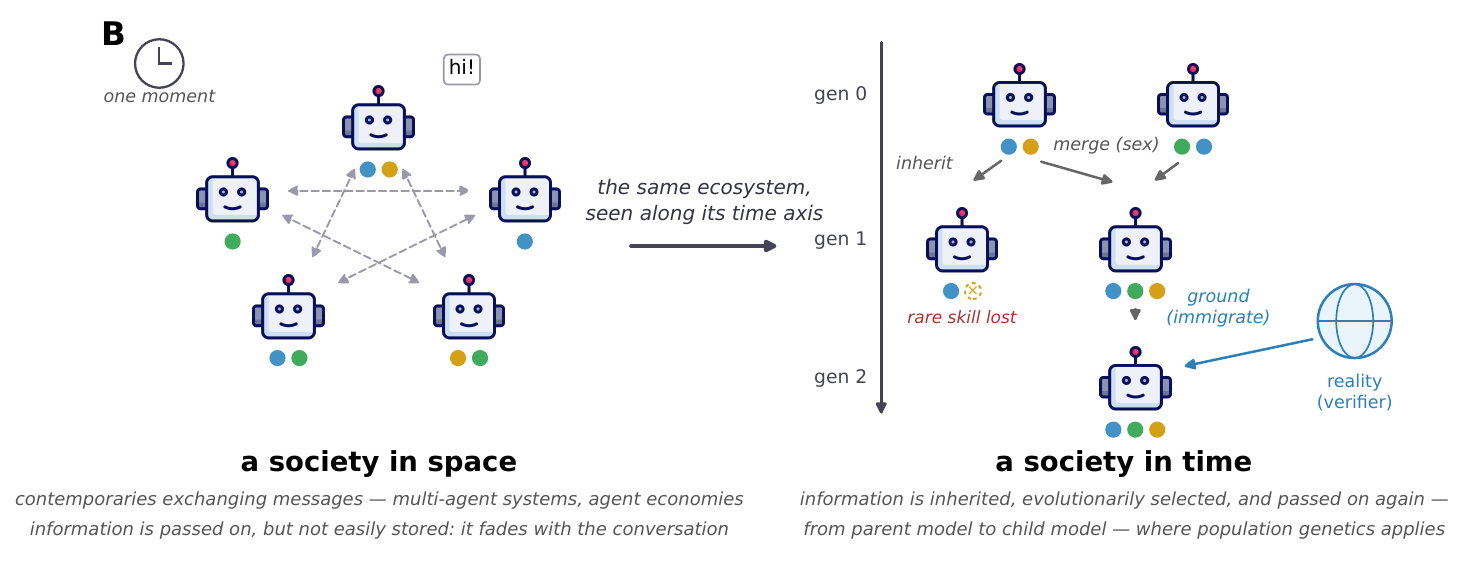}
\caption{A map of the study. (A) Each row is a biological mechanism the paper borrows, each column a level of realism at which it is tested: an inheritance model (an exact simulation of knowledge transmission, green), trained neural networks measured against exact oracles (light blue), and language models (dark blue). Filled cells name the experiments run at each level and, in the corner, the figure or table reporting them; grey cells were not run, either because the result is established in prior work (23, 34) or because that level adds no new test for that question. The inheritance model is the reference column: it sets the expectation the real-model experiments are read against. (B) The change of viewpoint the framework rests on. A group of models is usually pictured as a society in space, contemporaries exchanging messages. The couplings studied here run between generations: training on model output (inheritance), weight-space merging (recombination), and verified real data entering each generation (immigration from reality). That is a society in time, which is what population genetics describes. Dots are capabilities: the rare one (gold) is lost under single-parent inheritance, reassembled by merging complementary parents, and re-supplied by grounding.}\label{fig1}
\end{figure*}

\section*{Results}

\subsection*{Inheritance is Wright--Fisher drift, plus a measurable bias}

Knowledge is modelled as a probability distribution \(p_t\) over \(K\) discrete \emph{items}, each item receiving a probability and standing for a capability, a fact or a mode of behaviour. An item is the counterpart of an allele, and the \emph{capability} it stands for is the counterpart of that allele's phenotype. A reference distribution \(p^{*}\), the distribution of real data, which does not change over generations, gives each item its true frequency, and its rare tail (the items of lowest frequency) carries the knowledge most at risk. Following population genetics I call an item's frequency \(p_i\) its \emph{mass}, the probability that one sample drawn from the distribution is that item (the allele frequency of Table 1), and the mass of a set of items is the sum of their frequencies. One generation has a single parent and a single child (several parents are the subject of the merging section) and consists of three steps: draw \(n\) samples from the parent's distribution; optionally add \(m\) samples drawn from \(p^{*}\) itself, standing for real data that have passed a verifier (\emph{grounding}, with grounding fraction \(g = m/(n+m)\)); and fit the child's distribution to the pooled \(n + m\) samples (the \emph{refit}, which in the minimal model is simply the observed frequencies). The resampling step coincides exactly with the Wright--Fisher process, population genetics' canonical model of neutral evolution, in which each generation is a random sample of size \(n\) from the last. In this \emph{inheritance model} the Wright--Fisher ``population'' is the sample a child is trained on and its ``individuals'' are the \(n + m\) samples, so it is a model of a learner. Diversity throughout is \emph{heterozygosity}, \(H = 1 - \sum  p_i^{2}\), the probability that two items sampled independently from the distribution differ (high when the mass is spread over many items, zero when one item holds it all). The simulator reproduces three closed forms of the process to within 0.5\% of the analytic value (Methods): (i) the heterozygosity decay under drift alone, \(\mathrm{E}[H_t] = H_0(1 - 1/n)^t\); (ii) the stationary diversity under real data, written in the next subsection; and (iii) the expected fraction of rare items held by at least one of several parents, which rises with the number of parents and falls as the parents grow alike (for \(N\) parents that each hold an item with probability \(q\), with holdings correlated by \(\rho \), it is \(\rho q + (1 - \rho )(1 - (1 - q)^N)\)), used in the merging section.

Trained networks are not exact copiers, because they add approximation error, optimisation noise and their own inductive bias to the resampling step, so before using Wright--Fisher as a reference I measured how far real learners depart from it. Run through the same generational loop against an exact oracle, they departed in opposite directions (Fig. S2). The sequence generators (a recurrent and a feedforward network) \emph{smooth}, spreading probability onto items they have never seen, and so collapse more slowly than drift predicts while keeping spurious variants alive. The image autoencoder \emph{sharpens}, concentrating probability on its commonest modes, and so collapses faster (Fig. 2A; the comparison with drift in Fig. S2). Both departures are reproduced by adding one knob to the copying step, a mutation rate toward a prior for smoothing or a temperature for sharpening (Fig. S2). A real learner can therefore be faithfully modelled as Wright--Fisher plus a signed, measurable bias (Figs. 2 and S1).

In biological terms, retraining a child on a single parent corresponds to \emph{asexual reproduction}. In a population that never recombines, a loss that reaches every individual is permanent, because no copy remains from which to rebuild. Each such loss clicks the population's repertoire one notch down, and the notches turn only one way. Population genetics knows this mechanism as \emph{Muller's ratchet} (30), and model collapse has the same irreversible arm. The inheritance model shows the trap in its commonest form: a population that adopts its own collapsed output as its new reference never recovers the items it had lost, whatever real data it is fed afterwards (Fig. S3). Remedies must therefore act while copies still survive somewhere in the population.

\begin{table*}[tp]
\caption*{\textbf{Table 1.} The dictionary. Each biological term is introduced in the section that develops it. The support column references the evidence. ``Closed form'' means derived in the inheritance model and verified against simulation; ``empirical'' means measured in a trained system; ``hypothesis'' means stated with a falsifier and untested.}
\centering\footnotesize
\begin{tabular}{p{\dimexpr(\textwidth-6\tabcolsep)/3\relax} p{\dimexpr(\textwidth-6\tabcolsep)/3\relax} p{\dimexpr(\textwidth-6\tabcolsep)/3\relax}}
\hline
Population genetics & Model populations & Support \\ \hline
Genetic drift in a finite population & Training on finite samples of model output & Closed form (Fig. 2B); collapse measured (Fig. 2A); the identification is prior work (23--27) \\[3pt]
Immigration from a fixed source & Grounding with verified real data & Closed-form equilibrium and per-item floor (Fig. 2B); sign confirmed in trained nets (Fig. 2A); stationarity and stability under fresh data (31, 32); fraction reported (23); verification as the gate (33); accumulation regime (34); conservation analogue (35) \\[3pt]
Muller's ratchet (asexual decay) & Irreversible arm of model collapse & The irreversibility is reproduced in the inheritance model (Fig. S3); the mutational mechanism of the ratchet is not modelled (30) \\[3pt]
Recombination / sexual reproduction & Model merging & Fig. 3B--C: merging beats blending wherever the weight-average scores well below the best parent, and blending suffices where it does not; that merges can beat parents is established (36) \\[3pt]
Fisher--Muller effect & Merged specialists exceed every parent & Fig. 3B; inheritance-model expectation (Fig. S9); classical theory (37, 38) \\[3pt]
Outbreeding depression under epistasis & Merging entangled skills harms offspring & Inheritance model only (Fig. S10), reproducing (39, 40); hypothesis at LLM scale \\[3pt]
Mating systems / population structure & Who merges with whom (breadth of the parent pool) & Inheritance model only (Fig. S13), reproducing (41); hypothesis for real populations \\[3pt]
Reproductive isolation (Bateson--Dobzhansky--Muller incompatibilities) & Merge failure from functional conflict & Fig. 5A--D and SI Text S1, Proposition S2; emergent form not observed; classical theory (42, 43); alignment tools and known residuals (44--47) \\[3pt]
Seed bank (mating with a stored earlier generation) & Merging with one's own ancestor & Six-generation population (Results; SI Table S2): own-ancestor merge beat a contemporary in every seed; checkpoint averaging as a stabiliser (48, 49) \\[3pt]
Recombination modifier (a gene that sets how often other genes are shuffled) & A declinable merge: keeping the parent unchanged is scored as one candidate offspring & Fig. 4B--C (six generations, 3 seeds): a fixed early stop matched it, and declines tracked generation, not complementarity, once the two were decoupled. Modifier theory (50--52) is the motivating frame; its reduction-principle reading was not supported; gated and early-stopped merging in continual settings (53, 54) \\[3pt]
Selection on a fitness function & Verifier-anchored selection (``reality that can say no'') & Fig. 4D--F; diversity-preserving selection after (55), inheritance-model reference (Fig. S12) \\[3pt]
\hline\end{tabular}
\end{table*}

\subsection*{Grounding is immigration: a count, not a fraction}

Grounding, the mixing of verified real data into each generation's training sample, plays in the inheritance model the role that immigration plays in population genetics. An external source that does not itself change over generations (the real-data distribution \(p^{*}\)) supplies a fraction \(g\) of each generation's sample, and a population that would otherwise drift to fixation settles instead at a stationary diversity (33, 34, 56). I swept \(g\) from 0 to 0.4 across 100 independent lineages (Fig. 2B and Fig. S4) to separate two questions: how much real data hold aggregate diversity, and what happens to an individual rare item.

Part of the aggregate answer exists already: that a self-consuming loop fed fresh real data settles at a stationary state instead of collapsing was shown for generative models (31), a sufficient condition on the real fraction for stability has been proved (32), training on model output changes the scaling law itself (57), with any non-vanishing synthetic fraction enough to stop larger training sets from closing the gap to real data (58), and in the first collapse study retaining 10\% of the original data left only a minor degradation (23). These results establish that a grounded lineage stabilises below the real data without saying where. The inheritance model gives the level in closed form. With \(m\) real samples added to \(n\) inherited ones each generation, diversity settles at \(H_{\mathrm{eq}} = H^{*} \cdot  m(2n+m-1)/(n+2nm+m^{2})\), where \(H^{*}\) is the diversity of the source, and the simulator matches this to within 0.5\% (Fig. 2B). Two consequences follow that, without formalisation, the earlier results alone could not show. The first is that what holds diversity is the \emph{count} of real samples per generation, not their share of the training set. Whenever real samples are a minority (\(m \ll  n\)) the formula reduces to \(H_{\mathrm{eq}} \approx  H^{*} \cdot  2m/(2m+1)\) and \(n\) drops out: one real sample per generation keeps two thirds of the source's diversity and ten keep 95\%, however large the inherited sample is. The expression reproduces Wright's island model in haploid form. The shortfall \(1/(2m+1)\) is its fixation index \(F_{\mathrm{ST}}\) for a population receiving \(m\) migrants a generation, and the rule of thumb of conservation genetics is stated as \emph{one migrant per generation} (35), a count and not a fraction, because of the same cancellation of terms. In the tested setting (\(K = 1000\) items, \(n = 200\) inherited samples per generation, and a true distribution whose item frequencies fall off as a power law, a \emph{Zipf} distribution, the standard model of the long tail of natural data) 95\% of the source's diversity was kept from \(g \approx  0.05\) upward (Fig. S4), but that fraction is ten real samples divided by a training set of 200, and by the closed form the same ten samples would keep the same 95\% in a training set of any size, so the fraction needed shrinks as the training set grows. The second is that the curve is smooth. Diversity rises gradually with \(m\), there is no value at which a lineage switches from collapsing to safe, and the lineage never reaches the source (the shortfall is about \(1/(2m+1)\) at any budget, as the strong-collapse result requires; 58). A quoted real-data threshold therefore expresses a choice of how much diversity to retain, read off the curve, and not a property of the system. Interestingly, accumulating rather than replacing data, which lets the real share fall to a tenth of the pool after ten generations without collapse (34), and the replay ratios of continual learning point the same way. Optimal mixing ratios derived for squared-error regression are far higher (about 0.6; 59), because that objective weighs every sample equally where the question here is which items survive at all.

Aggregate diversity cannot say whether one particular rare item survives; that question has a direct answer. Call the number \(m\) of verified real samples added per generation the \emph{real-data budget}. Under unstratified sampling an item of frequency \(p\) appears in a batch of \(m\) real samples with probability \(1 - e^{-mp}\), so a budget of \(m \approx  1/p\) gives only a 63\% chance of seeing the item once per generation; an item that appears in one real sample in ten thousand needs a budget of about ten thousand real samples every generation. The budget is therefore set by the rarest item one refuses to lose, and it is a lower bound, because a single copy that does arrive enters a pool of \(n + m\) samples and can still be lost when the child is resampled from it (Fig. S4D, where the rarest items recover last). The rule is the immigration counterpart of the per-item extinction laws derived for closed loops (25, 60). It is the per-item form of the count-not-share rule of the previous paragraph, and the two together explain an observation (61), that the absolute count of real samples predicts collapse better than their proportion, since both depend on \(m\) and not on \(g\). The same arithmetic has been observed on the acquisition side, in pretraining itself: about 250 documents install a rare behaviour in models from 600 million to 13 billion parameters, although the larger models see twenty times more data, so the documents' share of the corpus falls twentyfold while their effect does not (62). One migrant per generation, 250 poisoned documents and \(m\cdot p \gtrsim  1\) are one rule read three times: what a population keeps, or acquires, of a rare item is set by the number of copies that reach it each generation, not by the size of everything else it is trained on. A fixed budget stretches further in two ways. Real data protect only the topics they cover, since when the 1,000 items are split into ten topics and the same budget is spent either on one topic or evenly over all ten, real data aimed at the topic keep about half of its rare items alive and real data spread over all topics keep 7\% (Fig. S5), so a capability is protected by real data about that capability, not by real data in general. And an item lost from one lineage can be recovered from another lineage that still holds it, which is the subject of the next section.

In the trained networks (the recurrent and feedforward generators on the synthetic universe, Fig. S6, and the convolutional VAE on MNIST, Fig. 2A and Fig. S7) grounding reduced collapse in every case. The trained networks depart from the exact model in two ways, both traceable to the bias measured above. First, the benefit of real data arrives more gradually. In the recurrent network the distance from the truth falls steadily over the whole range of \(g\) tested (Fig. S6B), whereas the inheritance model's diversity saturates within a few percent. Second, a smoothing learner defeats the usual measure of collapse. Such a network keeps assigning probability to items it was never trained on, so counting how many rare modes survive overstates its health (in the recurrent network the count is not even monotone in \(g\), Fig. S6D), and a network can retain every mode while holding the mass in the wrong proportions. For smoothing learners I therefore measure collapse by the Kullback--Leibler divergence from truth to model, the standard measure of how well a model covers a distribution, which penalises every region where the truth has mass and the model has little. On real images (Fig. 2A) ungrounded self-training collapsed a convolutional VAE from thirty modes to one within fifteen generations, while about 10\% real data held all thirty (Fig. S7). The autoencoder needed about 10\% real data where the inheritance model needed 5\%, and the difference is the cost of its sharpening bias. A learner that concentrates mass on its commonest modes loses rare ones faster than sampling alone would, and needs more real copies to hold them.

\begin{figure*}[p]\centering  % fig2
\includegraphics[width=\textwidth,height=0.56\textheight,keepaspectratio]{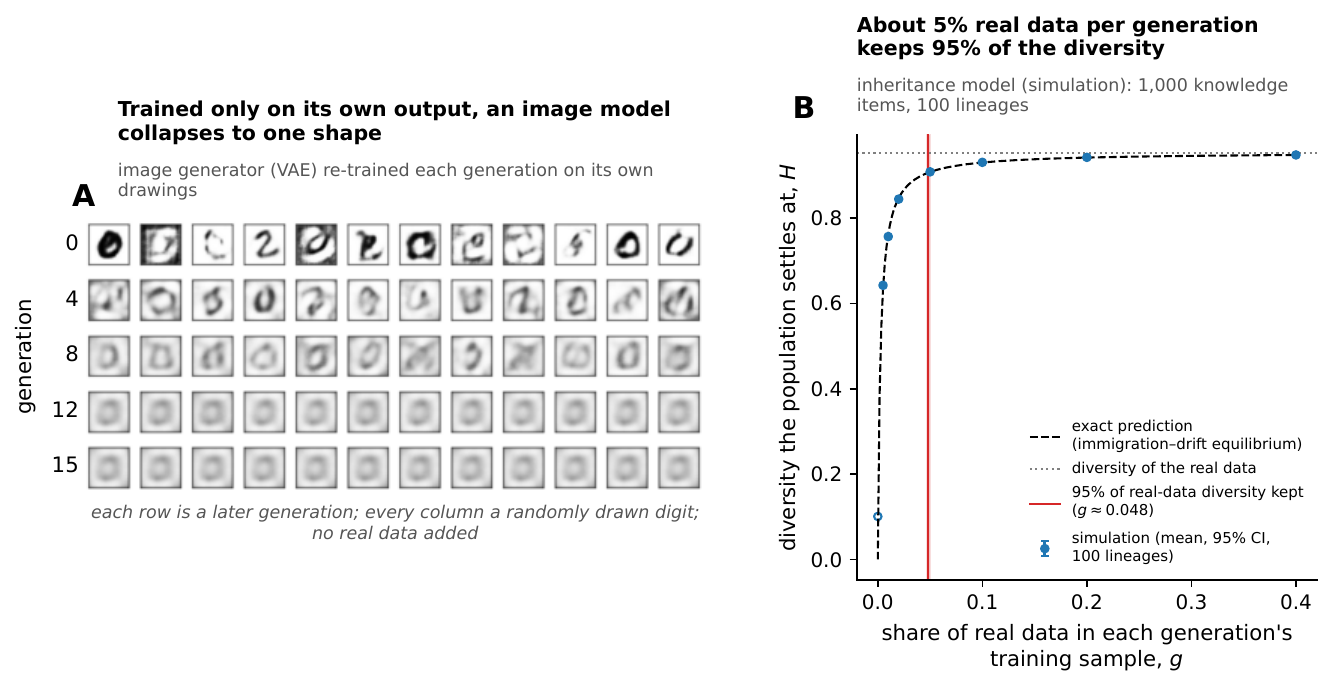}
\caption{The amount of real data that stops model collapse. (A) An image-generating network (a variational autoencoder) is trained on handwritten digits, then a fresh copy is trained only on the digits the previous one drew, for fifteen generations, with no real data added. Each row is a later generation (0, 4, 8, 12, 15) and each column a randomly chosen drawing. The thirty kinds of digit (ten digits $\times$ three stroke thicknesses, some kinds rare) collapse to one blurred shape; an independent classifier confirms that the number of kinds still drawn falls from 30 to 1, while adding 10\% real digits each generation keeps all 30 (Fig.~S7; 4 replicates). (B) The same question in the inheritance model, the exact simulation: 1,000 knowledge items, 200 samples drawn per generation, and a fraction $g$ of fresh real samples mixed in. Points are the diversity the population settles at after 500 generations (mean and 95\% CI over 100 lineages), the dashed line the exact prediction (the immigration--drift equilibrium), the dotted line the diversity of the real data themselves. The curve is smooth, so any threshold is a choice: the red line marks the $g$ at which 95\% of the real data's diversity is kept, about 0.05 (bootstrap CI shaded). The hollow point at $g = 0$ has not yet reached its equilibrium of zero. The trained image model needed about twice this fraction, because a trained network is not the exact copier the simulation assumes (Fig.~S2).}\label{fig2}
\end{figure*}

\subsection*{Averaging is blending inheritance; recombining is Fisher--Muller}

Refitting a child on the average of its parents' output distributions revives \emph{blending inheritance}, the pre-Mendelian view of heredity in which offspring are an average of their parents. Fleeming Jenkin's objection to Darwin (63) was that under blending a rare favourable variant is halved at every cross and swamped within a few generations, so selection could never establish it; particulate (Mendelian) inheritance, in which an allele passes intact or not at all, answered the objection, and blending was abandoned as a theory of heredity; Jenkin's own arithmetic was later shown to fail even under blending once recurrent mutation is allowed (64). I reason that, in machine learning, averaging does to a rare capability exactly what Jenkin said blending would do to a rare variant, and blending inheritance is therefore the right null model of merging. The same dilution has been reported in machine learning under three different names, without being recognised as one phenomenon: distilling onto an ensemble mean discards the members' diversity (65), averaging expert weights loses to routing among the same experts (66), and an update held by one of \(N\) parents is scaled by \(1/N\) in their soup (67). In the inheritance model the dilution is a conservation law: the expected mass of a rare item in the child is \(q\cdot p\) (its mass \(p\) in a parent that holds it, times the probability \(q\) that a parent holds it) whatever the number of parents, so averaging over more parents neither helps nor harms a rare item's expected share, and the proposition below says exactly when the same holds for its survival.

\textbf{Proposition (blending inheritance, rare-item regime).} Let each of \(N\) parents independently retain a rare item, which has mass \(p\) in a parent that retains it, and let the child draw \(n\) samples either from one parent chosen at random or from the mean of the \(N\) parents' distributions. The expected mass of the item in the child's sample is the same under both schemes. When the item is rare enough that even a parent holding it rarely contributes more than one copy to the child's sample (\(n\cdot p \ll  1\)), the probability that the item survives into the child is the same too: averaging over \(N\) parents makes the item \(N\) times more likely to be present in the mixture, and \(N\) times less frequent when it is, and the two factors cancel (proof in SI Text S2).

The proposition fixes the baseline against which any merging operator is judged, and it has two boundaries. For items common enough that the child usually sees several copies, averaging is safer than inheriting from one random parent, because the probability of losing an item is a convex function of its mass and averaging evens out which parent happened to hold it; the cancellation is a statement about rare items, which are the ones at risk. A \emph{union} operator, which keeps for each item the mass it has in the parent holding it most strongly (and therefore needs a verifier to say which parent that is), raises expected retention with every additional parent at every rarity tested (Fig. S8).

Neither scheme is what model merging does in practice. The two operators in use are \emph{weight averaging}, which averages the parents' parameters (a network is nonlinear in its weights, so averaging weights does not average outputs and the proposition applies only by analogy; but an update held by one of \(N\) parents is still scaled by \(1/N\) in the average (67), which is the dilution the proposition describes), and \emph{routing}, which keeps every specialist intact and sends each input to the specialist trained for it (68), the practical form of the union. I compared the two at two model sizes (0.5B and 7B parameters) on easy and on deliberately hard task families (Fig. 3C for 7B on the hard families; the other combinations in Supplementary Information, Table S2). On the hard families the weight average fell to the level of the best single specialist (0.41 for both, over three 7B seeds), because averaging dilutes each specialist's own skill, and routing among the intact specialists scored 0.50, ahead in every seed. On the easy families the 7B average already scored at ceiling on two of the three families (1.00 on both); those tasks were within reach of the base model itself, so specialisation contributed little, its dilution cost nothing, and the two operators were equivalent. A weak base (0.5B) showed the routing advantage even on the easy families. The variable that governs the gap is therefore the \emph{headroom}, the distance between what the weight average scores and what the specialists would jointly score if every input reached the right one. It is large wherever dilution has something to destroy (a weak base, or hard tasks at a strong one), and neither model size nor task difficulty alone predicts it. A second base lineage gave the same ordering with a larger margin (routing 0.33 against averaging 0.17 on the hard families at 1.7B, ahead in every seed, with the average below the best specialist in every seed; Fig. S16). Whether the gain scales quantitatively with the headroom is untested.

In an asexual population two useful variants that arise in different individuals can never meet in one descendant; the lineages carrying them compete, and one is lost. Recombination puts both into one offspring, which is why sexual populations adapt faster, an advantage known as the \emph{Fisher--Muller effect} (37, 38). Its counterpart here is that merging complementary specialists can yield a model better than any of them. In the multi-locus inheritance model, merged decorrelated specialists reach a combination of variants (a \emph{genotype}) that no parent held, while the best parent and the blended average plateau below (Fig. S9). Merges of three LoRA (69) specialists reproduced the signature, beating every parent overall (0.65 against 0.59 over five seeds at 0.5B; 0.87 against 0.81 over three seeds at 7B, in every seed), and on \emph{worst-family accuracy}, a model's score on the task family it handles least well, they were the only models competent everywhere, in every seed (Fig. 3B). The same protocol on an unrelated base lineage (SmolLM2-1.7B-Instruct: a different laboratory, architecture family and pretraining corpus) gave the same result in every one of five seeds (merge 0.66 against best specialist 0.61 overall; worst family 0.32 against 0.13; Fig. S16). That merges can exceed their parents is established for whole fine-tuned models (36) and for their fine-tuning deltas (70); the model contributes the condition under which it happens and the operator that realises it.

Blind recombination is not always safe. When the value of one variant depends on which other variants accompany it (\emph{epistasis} in genetics; entangled rather than independent skills in machine learning), the fitness landscape is rugged (71), and two well-adapted parents can produce offspring worse than either, because recombination breaks apart the combinations that made each parent work; the more entangled the traits, the lower the recombination rate that does best. Both results are long established in population genetics (39) and evolutionary computation (41) and are reproduced here only to fix reference values (Fig. S10). An engineered population has an option a natural one lacks: breed many candidate offspring and keep whichever a verifier scores highest. In the inheritance model this \emph{directed} recombination recovers the gain on every landscape where blind recombination loses it (Fig. S11), and in language models it beat the a-priori blend in every seed on hard tasks, including one seed where the blend failed catastrophically and selection was unaffected (Supplementary Information, Table S2).

\begin{figure*}[p]\centering  % fig3
\includegraphics[width=\textwidth,height=0.56\textheight,keepaspectratio]{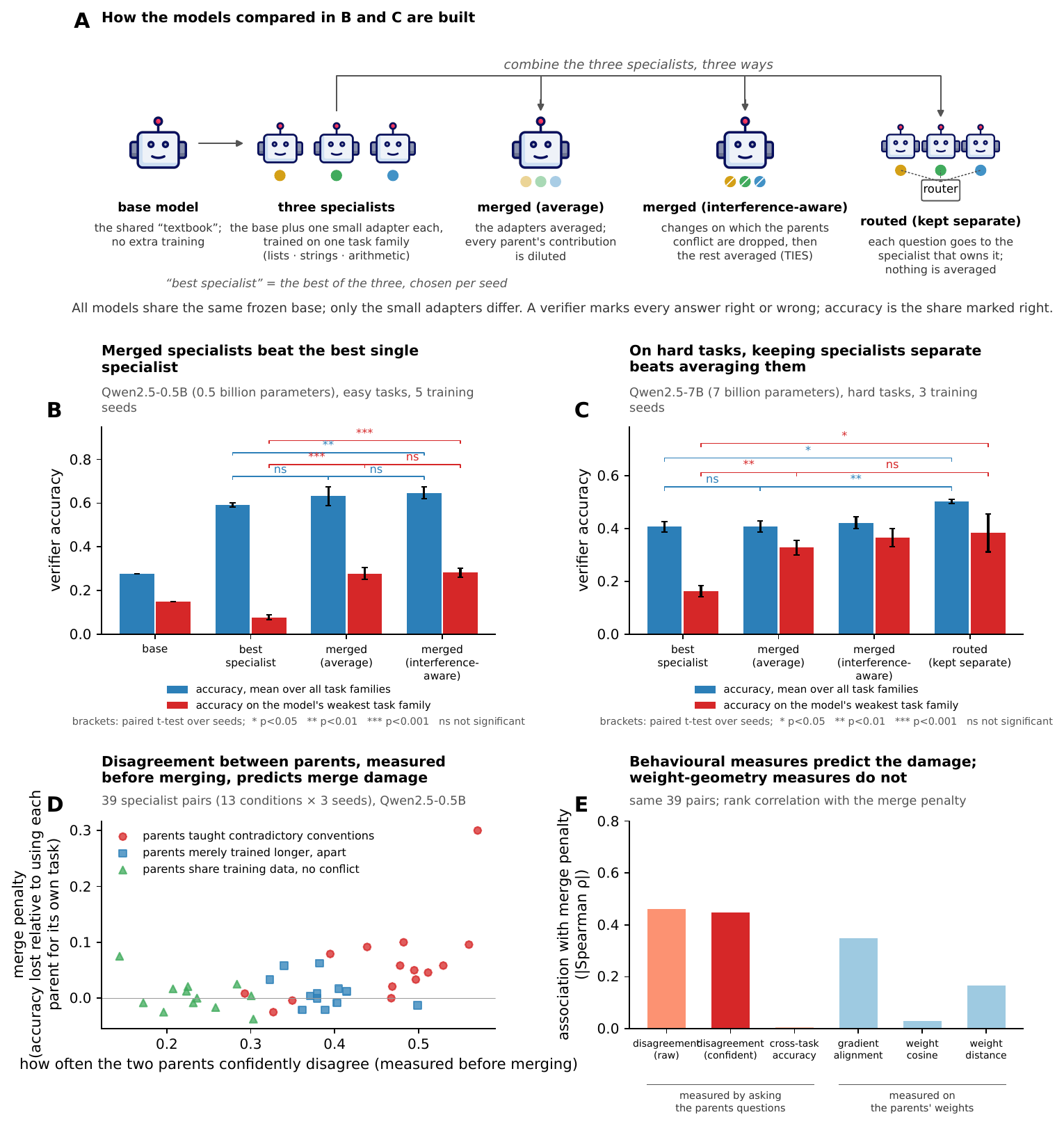}
\caption{Merging language-model specialists: when it helps, and predicting when it will hurt. All models are built from one frozen base (Qwen2.5) plus a LoRA adapter, a small set of extra weights trained on one family of tasks (list puzzles, string puzzles or arithmetic); a verifier marks every answer right or wrong, and accuracy is the share marked right on held-out questions. (A) The models compared: the base alone; three specialists (one adapter each); their merge by averaging the adapters; their merge after dropping the changes on which the parents conflict (TIES); and routing, which keeps the specialists separate and sends each question to the one that owns it. (B) Easy tasks, 0.5-billion-parameter base, five training seeds (fixed test sets; mean and 95\% CI). Both merges beat the best single specialist on the weakest task family (paired t-test over seeds, $p < 10^{-4}$), and the interference-aware merge beats it overall ($p = 0.006$; the plain average $p = 0.09$, ahead in 4 of 5 seeds); the two merges do not differ from each other. Only merged models are competent on every family. (C) Deliberately hard tasks, 7-billion-parameter base, three seeds. Averaging only matches the best specialist overall ($p = 0.96$) although it lifts the weakest family ($p = 0.009$); routing beats averaging overall ($p = 0.007$, ahead in every seed) and beats the best specialist on both measures ($p = 0.018$ and $0.014$). With three seeds, some comparisons that hold in every seed are not significant (ns). (D) Predicting merge damage before merging: 39 pairs of specialists built along three axes, parents taught contradictory conventions (red), parents merely trained longer on different tasks (blue), and parents sharing training data without conflict (green). The horizontal axis is how often the two parents confidently disagree when asked the same questions before merging; the vertical axis is the merge penalty, the accuracy the merged model loses relative to answering each task with the parent that owns it. Damage concentrates in the conflicting pairs. (E) Six pre-merge measures ranked by how strongly they track the penalty (absolute Spearman correlation): measures taken by asking the parents questions carry the signal, measures taken on the parents' weights do not; differences between individual predictors are not significant at this sample size (Table~S2).}\label{fig3}
\end{figure*}

\subsection*{Grounding, recombination and diversity each fail in their own way}

Grounding enters a population at two points. When a child is retrained on samples, as in the inheritance model above, grounding takes the form of \emph{grounded inheritance}, real samples added to the pooled sample the child is fit to. When agents are selected on a score, it takes the form of \emph{grounded evaluation}: an agent is scored partly against reality and partly against the population's own consensus (\(g\)\(\cdot\)true-fitness + (1\(-\)g)\(\cdot\)conformity). The consensus term models a population with no external check on its answers (in machine-learning terms, no verifier, no program or dataset able to mark an answer wrong), which can only score its members on how far they agree with one another; at \(g = 0\) agreement is all that is rewarded. To ask whether grounding, recombination and diversity contribute separately, I ran a four-arm ablation in the multi-locus inheritance model: a population of 60 agents, each a genotype of 12 loci, adapting on a rugged (NK) landscape for 80 generations (SI Methods M3), with one operator removed per arm (Fig. 4D--F). The full system (grounded evaluation, directed recombination, and diversity-preserving selection (55; its inheritance-model reference in Fig. S12)) approached the global optimum while keeping its specialists (Fig. 4D). Removing grounded evaluation converged the population confidently on an unfit consensus, the self-consumption failure (Fig. 4D and F). Removing recombination stranded it on local optima, and removing diversity converged it prematurely on a worse answer, draining diversity fastest (Fig. 4D and E). The arm without grounding fails by construction, since a rule that scores agreement will converge on agreement, but the other two removals fail in ways of their own, so under these conditions recombination and diversity are not substitutes for grounding or for each other. Magnitudes depend on the mutation, restart and selection schemes, which were not varied.

\subsection*{A multigenerational LLM society: the partner matters more than the mating}

Others have merged models repeatedly over several rounds, in two forms. Evolutionary merging holds a pool of parents fixed and recombines it again and again (5, 8, 9); over several generations the pool stagnates as its members grow alike (10). Continual merging folds a stream of independently trained experts into one running model (53, 54, 72, 73); in long streams it degrades unless merging is gated by similarity (54), and it can be stopped early at no cost once late experts add little (53). In neither form does a lineage learn a new skill by training between merges, so what happens to a composed capability when it is inherited, extended and recombined has not been measured. I therefore ran inheritance, recombination and immigration together in a population of language models across six generations on real datasets.

Three lineages start from one frozen base model (Qwen2.5, 1.5 billion parameters, untrained on the tasks; Fig. 4A). Each generation, every lineage acquires one new skill from six public datasets (natural-language inference (MNLI; 74), science questions (ARC-Easy; 75), commonsense completion (HellaSwag; 76), reading-comprehension spans (SQuAD; 77), yes/no questions (BoolQ; 78), pronoun resolution (WinoGrande; 79)), each scored by its own verifier, a program that marks an answer right or wrong. A skill lives in a \emph{LoRA adapter}, a small set of trainable weights added to the frozen base (the base a shared textbook, the adapter one specialist's margin notes). A child starts from its parent's adapter as it stands, trains it further on its own new skill, and passes the result on in turn, so what the parent learned in its lifetime reaches the child (the inheritance of acquired characters that Lamarck proposed and biology rejected, and that a weight file makes trivial). Each child's training set also contains a fixed number of examples from the skills its lineage learned in earlier generations (150, beside 300 new), so that new training does not overwrite old skills. This \emph{replay} is the standard remedy for forgetting in continual learning (28, 29), and in the terms of this paper it is grounding, real examples of the old skills entering each generation.

The six skills are scheduled as a Latin square, each lineage taking the same skills in a different rotated order, like three students working through one syllabus in different sequences. A partner therefore knows things a lineage lacks early on (\emph{complementarity}, the share of the partner's skills one lacks, is 1.0 at the first two generations) and nothing it lacks by the end (0.0 at the sixth). Complementarity is thus a swept variable, but it falls as generation number rises, so this curriculum alone cannot tell an effect of complementarity from an effect of the adapters' age; a second curriculum, below, breaks the collinearity. Merging averages two adapters at a mixing weight chosen on validation data (Methods). The arms are: never merge; always merge with a contemporary from another lineage (with verified or with self-generated replay); merge with one's own ancestor three generations back; and a \emph{declinable} merge, in which keeping the parent unchanged is scored as a candidate beside every merge and wins if none beats it. A control arm merges obligately through generation 2 and never afterwards (a \emph{forced stop}), the fixed schedule the declinable arm must be compared against. Lineages are never culled, so the population has inheritance, recombination and immigration of new skills but no differential reproduction. Each arm was run with three training seeds, and a lineage's outcome is its mean accuracy over the six skills.

Obligate recombination collapsed (Fig. 4B). The always-merge arm tracked the never-merge arm for three generations, then fell from 0.65 to 0.27, beginning when partner complementarity dropped below 0.8; its self-replay variant did the same (0.31), so replay was not what failed. The declinable arm neither collapsed nor won: it led at the start (0.68 against 0.60), was overtaken, and finished level with never merging (0.792 against 0.796; per-seed \(-\)0.03, +0.01, +0.01), while one model taught the curriculum alone reached 0.80 (with replay, forgetting was not a pressure recombination could relieve). In both non-obligate arms accuracy on the skills a lineage had been taught held near 0.78 and the first skill learned never eroded (0.85 \(\rightarrow\) 0.88); the obligate arm fell to 0.24 on those same skills.

These results show that the choice of partner mattered more than whether to merge. Merging with one's own ancestor three generations back, a partner that lacks the lineage's three most recent skills but shares every convention it holds, beat merging with a contemporary in every seed (0.66 against 0.27). The ancestor supplies complementarity in time: what it lacks is exactly what the lineage has since learned, and nothing it holds was learned differently. In population genetics a \emph{seed bank} plays this role, letting a population mate with its own stored past; in continual learning, averaging a model with its own earlier checkpoint is a known stabiliser (48, 49), and the matched comparison against a contemporary partner is what this population adds. In the declinable arm the share of proposed merges that were declined rose from 0.44 to 1.00 across the six generations (Fig. 4C), until the population had become the never-merge arm by its own choice. The forced-stop control finished level with it in every seed (0.793 against 0.792; per-seed differences in Supplementary Information, Table S2), so the declinable arm's outcome is explained by when it stopped and not by which merges it chose. A second curriculum, in which every lineage starts with the same skill so that complementarity begins at zero, peaks at the third generation (0.70) and returns to zero, produced the same rise in declines (0.44 \(\rightarrow\) 0.89). Pooled over both curricula with generation controlled, declines did not track complementarity (partial Spearman \(\rho\) = \(-\)0.07, 95\% CI \(-\)0.21 to 0.09, n = 36) but did track generation (partial \(\rho\) = 0.31).

Three things rise with generation in both curricula: the adapters' training age, the number of skills each holds, and the arrival in every lineage of the two families whose answer conventions conflict (yes/no against 1/\penalty5000{}2). Two further curricula moved only the third, delivering the conflicting pair in generations 1--2 of every lineage or in generations 5--6, with the four compatible families filling the rest in rotated orders so that age and skill count rise identically in both (Fig. S14). Neither the decline curve nor the collapse moved with the conflict. Declines rose with generation on the same schedule in both (0.56 \(\rightarrow\) 0.78 and 0.44 \(\rightarrow\) 0.89) and, with generation controlled, tracked generation (partial \(\rho\) = 0.45) and not the presence of conflict (partial \(\rho\) = \(-\)0.09, 95\% CI \(-\)0.45 to 0.15, n = 36). The obligate arm collapsed in both (final accuracy 0.28 and 0.39 against 0.80 and 0.78 for never merging, in every seed); the conflict-early population dipped when the pair arrived, recovered by generation 3 and collapsed from generation 5, while the conflict-late population collapsed from generation 4 with its conflicting pair still to come. What the four curricula leave confounded is adapter age with skill count, which rise together by construction.

A skill whose answer convention conflicts with nothing a lineage holds occupies a \emph{new locus}, a new position in the genome filled without displacing anything, and lineages accumulate loci freely (six here; half a million facts in a lifelong-editing benchmark that averages a fresh adapter per period into the accumulated one; 80). Two skills demanding different conventions for the same kind of question (``yes/no'' against ``1/\penalty5000{}2'' for a two-way choice) are \emph{alternative alleles at one locus}, and a model, like a chromosome, carries one. Where conventions disagree a merged child must err against at least one parent (SI Text S1, Proposition S2), and a lineage obliged to merge pays that error every generation on every pair of conflicting conventions until the errors accumulate into collapse. In the Latin-square curriculum the collapse began at the generation when partners stopped bringing skills a lineage lacked and started bringing conventions that clashed with the ones it held, but moving the clash by four generations did not move the collapse. Conflicting conventions therefore set the size of each merge's error, and something that grows with generation sets when the errors stop being repaired. Single models show the same divide: non-contradictory updates integrate safely while contradictory ones corrupt unrelated knowledge (81), and disjoint tasks make forgetting eliminable where conflicting overlap imposes a floor (82).

The declinable merge was designed as a \emph{recombination modifier}, in genetics a gene that sets how often other genes are shuffled between parents. Modifier theory holds that recombination is favoured when it assembles complementary alleles from different parents and disfavoured when it breaks combinations that already work (39, 50, 51), and that when shuffling gains nothing the \emph{reduction principle} drives its rate to zero (52); on that reading the declinable merge should have switched itself off as partners stopped being complementary. The controls do not support that reading here, since acceptance fell with generation whether or not partners were complementary and a fixed schedule reproduced the outcome. What the population establishes is narrower: one bit of selection on each recombination event, or a fixed early stop, avoids the collapse of obligate merging at no cost against never merging, and the declinable version does so without knowing in advance when to stop. The result was obtained under six generations, a single base model and replay throughout, none of which was varied, and with no differential reproduction. The Fisher--Muller argument predicts that selection is what turns recombination's early lead into a level advantage, because a lineage that assembles the skills first leaves more descendants. Adding truncation selection (after every generation the lowest-scoring lineage is re-founded from the highest, keeping its own place in the curriculum) did not bear this out (Fig. S15). Selection lifted the population mean early, but the final levels converged (with selection, never merging reached 0.804 and the declinable merge 0.793, below in every seed by 0.011 \(\pm\) 0.003, against 0.796 and 0.792 without it), and recombination's early lead was the same with and without selection and gone by generation 5 in both. Under a curriculum that delivers every skill to every lineage the ceiling is what one adapter can hold (0.80 for the single model taught the whole syllabus), and sex and selection each reach it sooner without raising it.

\begin{figure*}[p]\centering  % fig4
\includegraphics[width=\textwidth,height=0.56\textheight,keepaspectratio]{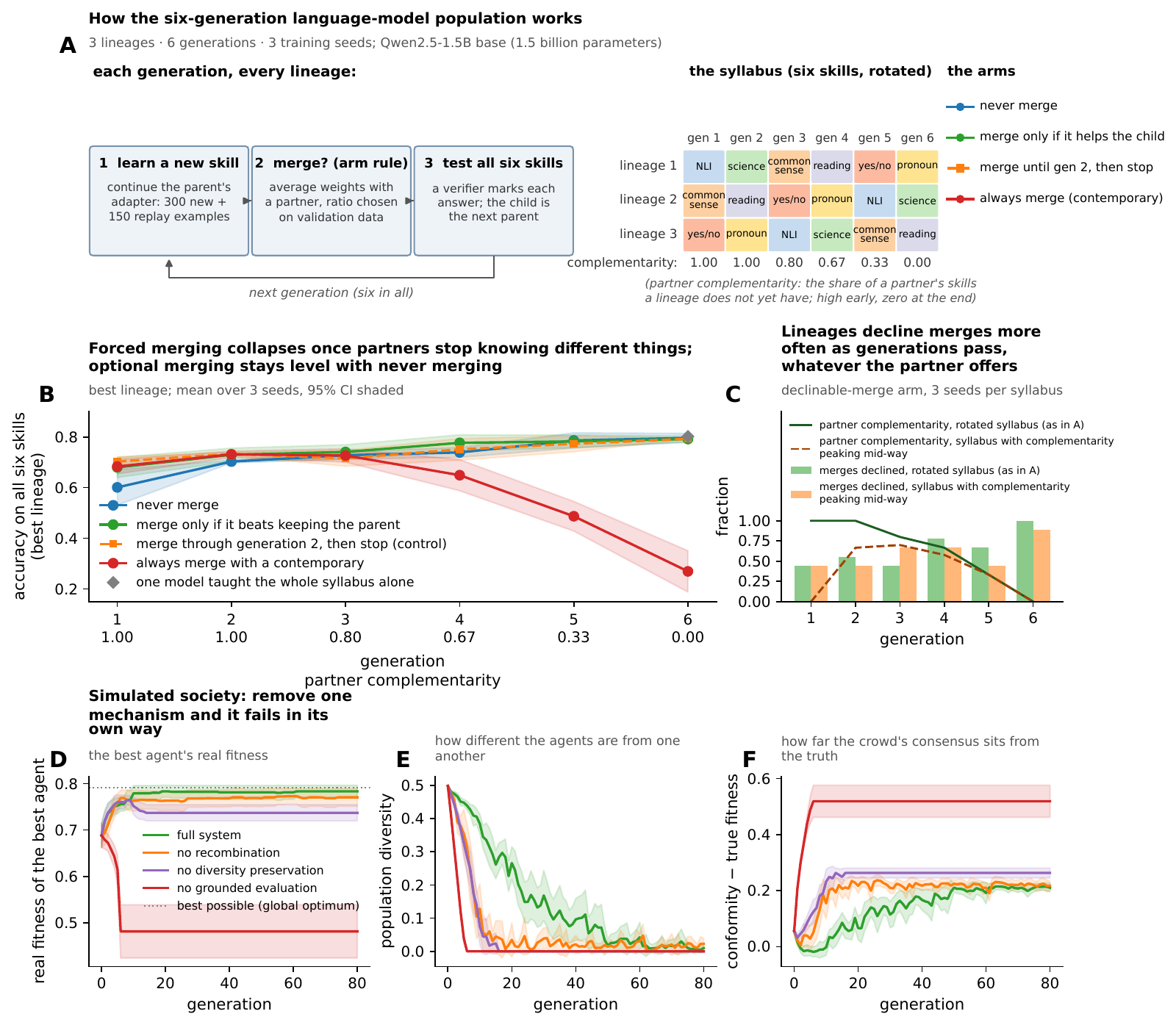}
\caption{A population of language models over six generations. (A) The set-up. Three lineages start from one frozen 1.5-billion-parameter base (Qwen2.5-1.5B). Each generation, every lineage learns one new skill from a public dataset by continuing to train its parent's adapter (300 new examples plus 150 replayed from earlier skills), may merge with a partner according to its arm's rule (weights averaged at a ratio chosen on validation data), and is tested on all six skills by a verifier; the child becomes the next parent. The six skills are taken in rotated order, so a partner knows things a lineage lacks early on (complementarity 1.0) and nothing it lacks by the end (0.0). Three training seeds. (B) Accuracy over all six skills of the best lineage (mean and 95\% CI). Never merging and merging only when it beats keeping the parent finish level (0.80 and 0.79); merging with a contemporary every generation collapses to 0.27, beginning when partners stop being complementary; a control that merges through generation 2 and then stops (dashed) matches the declinable arm in every seed, and a single model taught the whole syllabus alone (diamond) matches the population. (C) How often the declinable lineages refused a merge (bars) against partner complementarity (lines), under the rotated syllabus and under a second syllabus in which complementarity is zero at the start, peaks mid-way and returns to zero. Refusals rise with generation under both; with generation held fixed they do not track complementarity (partial Spearman $\rho = -0.07$, 95\% CI $-0.21$ to $0.09$, $n = 36$). (D--F) The simulation that motivated the design: 60 agents evolving on a rugged fitness landscape with all four mechanisms (grounded evaluation, recombination, diversity preservation, mutation) and one removed per arm (12 replicates; mean and 95\% CI). Removing grounded evaluation, so that agents are scored on agreement with the crowd instead of on the truth, collapses the population onto a confident but wrong consensus (D, F); removing recombination or diversity preservation strands it below the optimum (D) and drains diversity fastest (E). Each removal fails in its own way.}\label{fig4}
\end{figure*}

\subsection*{Conflicting conventions, not divergence, cause reproductive isolation}

Recombination presupposes compatible parents. In biology, lineages pushed far enough apart become separate species (\emph{reproductive isolation}) through Bateson--Dobzhansky--Muller incompatibilities (42, 43), changes harmless on their own genetic background but deleterious in combination. This is the mechanism behind the mule's sterility, in which two genomes that each work cannot run in the same cell. A merged model plays the part of that hybrid. In the inheritance model of the process (Fig. 5 E and F) hybrid fitness stays at the parents' level while the lineages remain compatible and then falls below the ancestor, sooner the more incompatibilities the genomes carry, and Orr showed that the number of such incompatibilities grows with the square of divergence (43). Whether a growing number of conflicts produces a fall in performance in a trained network is the question the simulation cannot answer.

In trained networks the claim must survive a known alternative. Two networks trained separately can differ in their weights for a trivial reason: the hidden units of a network can be renumbered, and in a ReLU network each unit's incoming weights can be scaled up and its outgoing weights scaled down by the same factor, without changing what the network computes. Two networks that compute similar functions can therefore lie far apart in weight space, and averaging them gives a poor model, a \emph{coordinate barrier}. Merge barriers between independently trained networks are famously of this kind, removable by re-aligning hidden units (44) and renormalising their activations (46) before averaging, and richer symmetry groups remove more (83). A residual that alignment does not remove is also known: networks trained on different tasks keep a barrier after permutation (47), and experts diverged far from a shared base keep one with symmetries accounted for (45). What has not been asked is what the residual measures, divergence as such or conflict in what the networks compute. To separate the two I aligned pairs of networks under permutation matching combined with exact per-unit rescaling (the complete unit symmetry group of plain ReLU MLPs; 44, 46) and measured the barrier before and after (Fig. 5 A and B). Two networks trained from different initialisations on the \emph{same} task have a barrier the alignment removes almost entirely (residual \(\approx\) 0.001, the aligned merge performing at parent level): their barrier was coordinate mismatch. Two networks trained on \emph{conflicting} label maps (the same inputs, with a fraction of the classes relabelled) have a barrier the alignment leaves unchanged (0.502 \(\rightarrow\) 0.497), and the merged model is functionally dead. The aligner is validated only on a special case (exact recovery of a permuted-and-rescaled copy of a network), so the share of the barrier it removes is a lower bound on the removable share, and the residual an upper bound. Sweeping the fraction of classes in conflict traces the fall in hybrid fitness from 0.97 to 0.03. A single model cannot answer one prompt in two ways, so where the parents' conventions contradict the merged model must err against at least one of them, whatever the training (SI Text S1, Proposition S2). The population view adds where the cliff sits. It moves with the share of shared inputs on which the parents' conventions contradict (Fig. 5B), and in a population that share grows whenever lineages adopt conventions independently.

The sharpest test asks whether isolation can emerge without any conflicting signal, as a true Bateson--Dobzhansky--Muller incompatibility would, each lineage's changes being harmless on their own. I diverged pairs of networks with no conflict anywhere, giving each the classes the other lacked and a different input convention, for up to 6.4 times the base training. No isolation emerged (residual 0.000 throughout). The merge instead rescued the two specialists: each had forgotten the other's classes and scored about 0.50 alone, while their weight average scored 0.955 at every divergence tested, the strongest Fisher--Muller effect in the paper. Language models gave the same double result in each of three training seeds (Fig. 5 C and D). Conflicting conventions broke the merged model on the conflicted function (0.02, 0.12 and 0.16 at full conflict, against 0.23--0.25 for either parent), while disjoint skills trained on the same budget merged unharmed, and over-training disjoint specialists from 1 to 12 epochs (cf. the expert-duration effect; 84) produced no isolation, the merge improving in every seed (0.76 \(\rightarrow\) 0.95 on the parents' private tasks). Longer expert training has been reported to harm merging by plain averaging (84, 85) and deepening specialisation to lower feature similarity between experts (68); in the regimes tested here neither produced isolation without conflict (a complementary-class merge rescued by alignment had been seen before on label-skewed splits; 44). In every tier, isolation had to be provoked by functional conflict, and specialisation alone did not speciate. What breaks merging is conflicting conventions on shared circuitry, not divergence as such. This is the cost the obligate-merge arm of the six-generation population paid from its fourth generation onward, once its partners held skills it had already learned under conventions of its own (Fig. 4B).

\begin{figure*}[p]\centering  % fig5
\includegraphics[width=\textwidth,height=0.56\textheight,keepaspectratio]{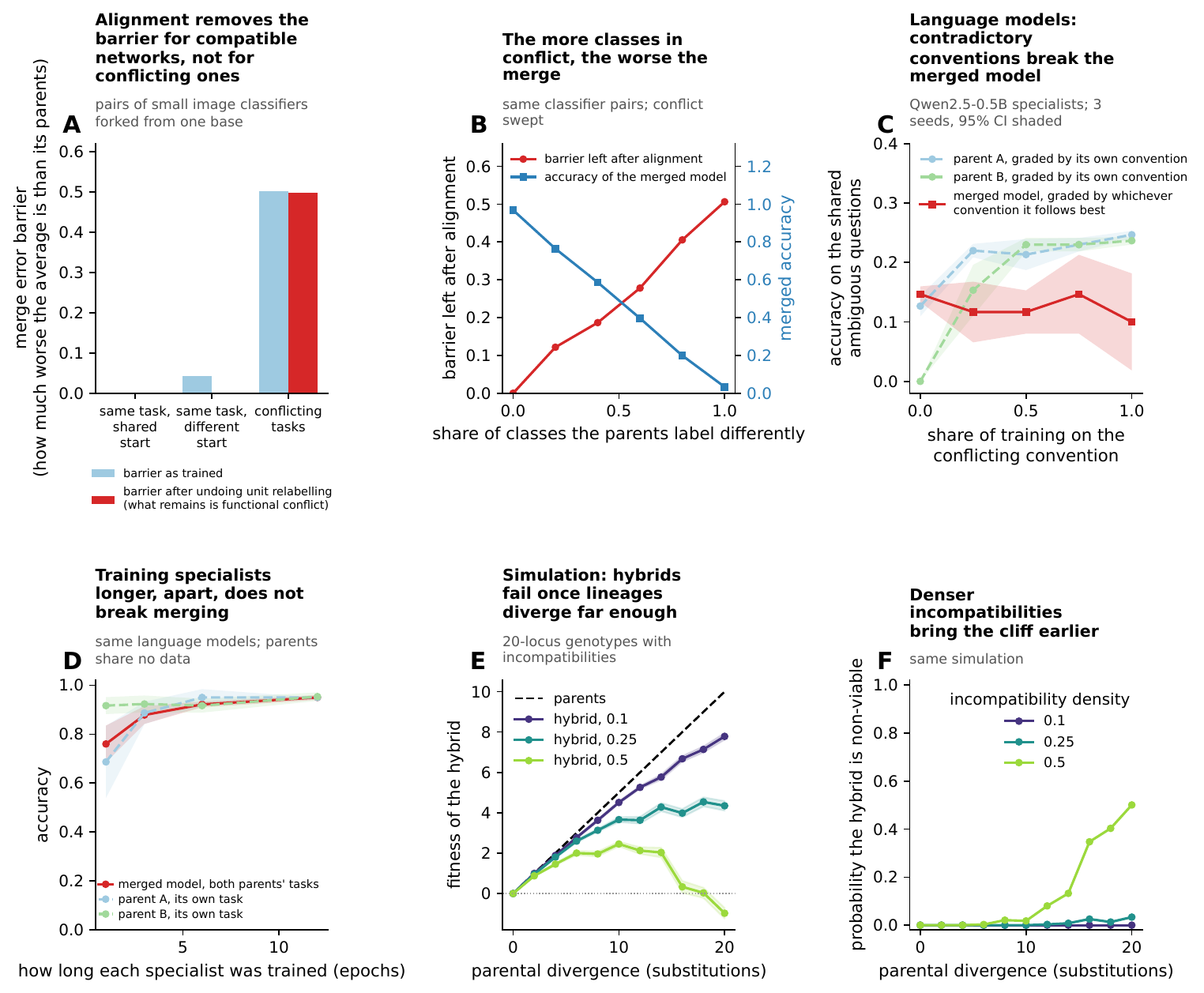}
\caption{Model speciation: when two lineages can no longer merge. (A, B) Small image classifiers (multilayer perceptrons) forked from one trained base. Two networks that compute the same function can still differ in their weights, because hidden units can be renumbered and rescaled without changing the output; alignment undoes this before averaging. The merge error barrier is how much worse the average of two networks is than the networks themselves. (A) Two copies trained from different random starts on the same task have a barrier that alignment removes almost entirely (0.04 to 0.001); two trained on conflicting labels (the same images, some classes relabelled) keep theirs (0.50), and their average is useless (3 replicates). (B) Sweeping the share of classes in conflict moves the merged model's accuracy from 0.97 to 0.03. (C) Language models: two specialists share a set of ambiguous questions (``sort this list'', direction unstated) and are taught opposite conventions. As the share of conflicting training grows, each parent stays good under its own convention while the merged model falls below both, in all three seeds (95\% CI shaded). (D) The control: specialists trained longer and longer on different tasks, with no conflict, merge better, not worse, in every seed. (E, F) The simulation: 20-position genotypes carrying incompatibilities of the Bateson--Dobzhansky--Muller kind. Hybrid fitness tracks the parents while lineages are compatible, then crashes, sooner the denser the incompatibilities (E), and the probability of a non-viable hybrid rises with divergence (F). What breaks merging is conflicting conventions on shared machinery, not distance or specialisation as such.}\label{fig5}
\end{figure*}

\subsection*{The parents' disagreement predicts hybrid load}

If functional conflict is what breaks a merge, measuring it on the parents should forecast the damage before any merge is made. I tested this on thirty-nine pairs of LoRA specialists (13 training conditions \(\times\) 3 seeds), built so that three properties of a pair vary independently of one another (Fig. 3D): \emph{conflict} (the parents answer the same prompts under contradictory conventions, with their private training budgets held fixed), \emph{compatible overlap} (the parents are trained on the same prompts under the same convention, so they share data and volume without conflict), and \emph{duration} (the parents are trained longer on disjoint tasks, so their weights diverge with no conflict at all).

Six quantities were computed on each pair before merging. Two are functional, obtained by putting the same probe questions to both parents (probes drawn without knowledge of where the conflict lies): the fraction of probes on which the parents answer differently (\emph{raw disagreement}), and the fraction on which they answer differently and both confidently (\emph{confidence-weighted conflict}, proposed here as the better proxy for merge-relevant interaction, because raw disagreement also counts the harmless case in which one parent is merely ignorant). Three describe the geometry of the parents' weight changes: the cosine similarity and the distance between the two LoRA updates, and the alignment of the two tasks' gradients at the shared base (86). The sixth is a baseline, each parent's accuracy on the other's task. The outcome is the \emph{merge penalty}, how far the merged model falls short of the accuracy the pair would reach if each task were answered by the parent trained on it. In population genetics that shortfall is \emph{hybrid load}, the fitness a hybrid loses relative to what its parents' genes could jointly supply.

Functional disagreement measured before merging predicted the merge penalty (Fig. 3 D and E). Its rank correlation with the penalty was \(\rho\) = +0.45 (+0.46 for the confidence-weighted variant), with a 95\% confidence interval excluding zero (bootstrapped over conditions, because the three seeds of one condition are not independent), and it kept \(\rho\) \(\approx\) 0.35--0.40 when each condition in turn was held out and predicted from the rest. The cosine and the distance between LoRA updates showed no detectable association, and gradient alignment carried intermediate signal. The direction agrees with three recent reports: hidden-state distance between parents tracks merging loss where four parameter-space metrics, cosine among them, do not (87); global cosine, sign conflict and subspace overlap miss functional interference between task vectors (88); and gradient distance outpredicts task-vector cosine in vision (86). Those studies are correlational or in-sample; the design here holds conditions out and adds the control below. At this sample size the differences between predictors are not individually significant, only these baselines were tested, and three seeds leave substantial uncertainty about generalisation, though the functional measures led within every seed taken alone (Supplementary Information, Table S2).

The compatible-overlap control produced a finding of its own. In an initial grid that varied only conflict and duration, the best predictor was the cosine between LoRA updates (\(\rho\) = +0.60). Parents trained on the same prompts have aligned weight changes and also merge worse, so the cosine was reading shared training data, not incompatibility: adding pairs that share prompts without conflicting collapsed its correlation to +0.03. Any merge predictor validated on a grid in which conflict and shared data vary together inherits this artefact. I know of no study that has controlled for it, and it bears on the merge-prediction literature (86--88) independently of the biology. One prediction failed. Confidence weighting did not beat raw disagreement as a rank predictor, so the evidence supports functional disagreement in general and not the incompatibility-specific refinement. Headline quantitative results, with sample sizes and uncertainty, are collected in Supplementary Information, Table S2.

\section*{Discussion}

What did these experiments find that matters? With generative AI, models can be trained on material created by other models, or built by averaging the weights of existing peers. These practices create vertical transmission, in which information is passed from one generation of models to the next. Biology has spent a century describing this very flow of information with population genetics, and here I show that the theory transfers almost wholesale across fields.

The first result treats immigration as a solution to model collapse and estimates how much real, human-made data a line of models needs in order not to degrade. When each generation learns only from the one before, rare knowledge disappears by chance, generation after generation, in the way that rare surnames vanish from small villages. It's known that adding real data to each generation slows this down (I call this grounding), but in what amount? The answer, surprisingly, is a count, not a percentage. What protects a rare skill is the number of real examples of it that reach each generation, and the size of everything else in the training set makes no difference. Ten real examples per generation keep 95\% of the variety, whether the training set holds two hundred items or two million (Fig. 2B, and the closed form behind it). Conservation biologists know this rule as ``one migrant per generation'' (35), and machine learning has met it twice before without recognising it: about 250 documents are enough to teach a rare behaviour to models from 600 million to 13 billion parameters (a technique called ``poisoning''; 62), and the rates at which continual-learning systems replay old data, 1\% in one setting and up to 25\% in another (89, 90), look inconsistent as percentages and agree once translated into numbers of examples per step. The practical consequence is a budget, set by the rarest skill one refuses to lose: a skill appearing in one real example in ten thousand needs about ten thousand of them every generation.

The second result describes, in population genetics terms, what happens when two models are combined by averaging their weights. Biologists first met this problem in 1867, when Fleeming Jenkin objected to Darwin that if offspring were an average of their parents, any rare favourable variant would be halved at every cross and swamped within a few generations (63). Mendel's discovery that traits pass on as discrete units answered him. In machine learning, Jenkin's objection revives: averaging works well when both parents are already good at everything, but fails when a skill is held by only one parent, because the average gives that skill half the weight. Three results reported under three different names in machine learning (65--67) strengthen the analogy. In the merges tested here, on hard tasks, averaging did no better than the best single specialist and sometimes worse, whereas keeping the specialists intact and sending each question to the one trained for it (routing) beat averaging in every seed and beat the best specialist overall. That routing can beat averaging is already on record (66); what the population view adds is when it will, since the gap is set by the headroom between what the average scores and what the specialists jointly could. The remedy is to stop averaging wherever the average falls short of the best parent, and to route, or to test several candidate merges and keep the best, instead.

The third result went further than I expected. I wanted to know what makes two models incompatible, so that a merge between them fails. That weight-space distance is an imperfect guide was already suspected (86--88), and much of the barrier between independently trained networks is known to be coordinate symmetry that alignment removes (44, 46). What I did not expect was that divergence would carry no signal at all, and that more of it would help. Two models that had learned the same skill in separate training runs had almost unrelated weight changes, and yet merged perfectly once their hidden units were re-aligned. Two models trained apart for 6.4 times longer than their original training merged as well as ever, and language models trained twelve times longer than usual merged better, against reports that longer expert training harms merging (84, 85). What broke merging, every time, was something else: the two models had been taught to answer the same kind of question in two different ways, for instance ``yes/no'' against ``1/\penalty5000{}2'' for a two-way choice. The unit of inheritance in a model population is therefore a convention, a way of answering a class of questions, more than a body of factual knowledge or a set of weights. Most differences in weights are harmless, much as most differences between two genomes are. The immediate use is a test: to predict whether a merge will fail, ask both parents the same questions and count how often they disagree. That count predicted the damage a merge would do, where distance in weight space did not, in a design that controlled for shared training data, a control the earlier studies of merge prediction lacked (86--88): on a grid that varied only conflict and duration, weight cosine looked like the best predictor of all (\(\rho\) = +0.60), and adding pairs that share training data without conflicting collapsed it to +0.03.

The same idea was tested directly. Three lineages of language models each learned a new skill every generation, kept in a small adapter on a frozen base, and could merge with one another over six generations. While partners still brought skills a lineage lacked, merging helped. Obligate merging then collapsed, from 0.65 to 0.27. Conflicting conventions decide how much each merge costs, but not when the collapse comes. Lineages that met the conflicting pair only in their last two generations collapsed from generation 4, before that pair arrived. What decides the timing is how many generations have passed. Two cheap policies avoided the collapse entirely: refusing a merge whenever the merged child scored below the unchanged parent, and merging instead with the lineage's own ancestor from three generations back, which shares every convention its descendant holds and lacks only what the descendant has learned since. Recombination made lineages learn faster without raising the level they finally reached, which is what one adapter can hold, and adding selection between lineages did not change that. This is what Fisher and Muller predict in exactly this situation, where every lineage was guaranteed every skill eventually: a gain in speed and nothing more. The one prediction that failed was that lineages would refuse merges in proportion to what a partner had to offer; refusals rose with generation regardless, and a fixed rule (merge for three generations, then stop) did as well.

All these findings sit beside continual learning (28, 29). Its remedies for forgetting are the operations of this paper applied to a single model. Replaying stored real data (91--93) is grounding; keeping new skills in small add-on modules and folding them into the main model later (94--98) is keeping lines separate and then merging them; and merging as a way of accumulating skills (72, 73, 80, 99, 100) fails there too when the skills contradict each other (81, 82). The observation that rare knowledge is learned last and forgotten first (101--103) is the loss of rare variants seen in a single model, though part of what looks like forgetting is a model failing to recognise which task it is being asked to do (104).

The open problem is the fitness function. In biology, fitness is imposed from outside: an organism either survives its environment or it does not, and no lineage votes on the criterion. A model population has no such guarantee, because the criterion is written by whoever runs it, and selection will improve whatever that criterion measures. Scored on agreement with its own members instead of against reality, the population here evolved to a confident consensus that was wrong, and stayed well below a population selected on true fitness (Fig. 4D--F). Verified real data therefore do two jobs that discussions of model collapse treat as one: they restore the variety lost to drift, and they supply the selective environment. Building that environment for large populations of models, through verification, replication and challenge, is the question this paper raises and does not answer. A second problem is epistasis: skills that depend on one another make blind merging dangerous, and merging then pays only when done sparingly and with the offspring screened (39--41; Figs. S10--S11, S13), but nobody yet has a way of measuring how entangled two real skills are. And drift reaches writing style, since models trained on model output lose the rare words and constructions that make a voice, and people who write with such models come to sound like one another (105--107).

What determines whether a lineage keeps a capability lies outside any single model: what enters each generation, which parents a child has, which merges are allowed, and what selection rewards. As a growing share of what each model learns comes from earlier models, that is where the behaviour of the whole system will be decided, and population genetics arrived there a century ago.

\section*{Materials and Methods}

Full procedures, parameters, and replicate counts are in Supplementary Information, Methods. Appendix 1 (\emph{The figures explained}) restates every main and supplementary figure with a legend that explains the machine-learning experiment behind it for readers from biology.

\textbf{Inheritance-model tier.} A NumPy/SciPy Wright--Fisher simulator over \(K\)-item distributions (knowledge as \(p_t\), Zipf-tailed truth \(p^{*}\), and drift--grounding--refit generations), extended with a learning kernel (a smoothing and a sharpening knob on the refit), multi-locus genotypes on additive and Kauffman NK landscapes, n-parent crossover, and finite-population loops. All parameters live in per-experiment YAML configs. Every run derives its randomness from one master seed (\texttt{SeedSequence.spawn}) and is bitwise reproducible. Scientific-validation tests assert the closed forms to within 0.5\% and run in CI alongside 151 further correctness tests.

\textbf{Neural tier.} Trained-network experiments realise the same abstractions against an exact oracle. Histogram, RNN, MLP and VAE generators run on a synthetic mode universe, where the histogram model reduces the harness exactly to the inheritance model (the bridge gate), and a convolutional VAE runs on MNIST with a frozen CNN oracle at 98.5\% mode accuracy (its confusion matrix is recorded as the measurement floor). Speciation experiments fork no-BatchNorm MLPs (784--512--512--10) from a shared base, weight-average them, and measure the error barrier along the straight line between the two weight vectors (the linear-mode-connectivity barrier) before and after alignment. Alignment composes deterministic Git Re-Basin permutation matching with exact per-unit scale canonicalisation, the unit symmetry group of this architecture class taken as the search space, and is gated by exact recovery of a permuted-and-rescaled copy. Control recovery does not establish global optimality.

\textbf{Language-model tier.} LoRA specialists (rank 16) on procedurally generated task families with an exact-match verifier, on frozen Qwen2.5-Instruct bases (0.5B on one 16 GB GPU; 7B on one L40S). Operators: weight-space merges via adapter arithmetic (the plain weight average, or soup, and TIES, which reconciles the sign of each parameter change across parents before averaging; 4), per-input routing, and Dirichlet-sampled offspring populations screened on held-out validation splits. The six-generation population uses the Qwen2.5-1.5B base model, six public datasets with per-family exact-match or execution verifiers, and rank-16 adapters continued from the parent adapter each generation (300 new and 150 replay examples, 3 epochs), merged over the weight grid \{0.5/\penalty5000{}0.5, 0.3/\penalty5000{}0.7, 0.7/\penalty5000{}0.3\}, the winning weight chosen on 20 validation items per family and reported on 60 held-out test items so that selecting the weight cannot inflate the reported accuracy, with the unchanged parent as a further candidate in the declinable arm; three training seeds. Multi-seed protocols fix the test sets and vary the training seed. The predictive test computes all predictors pre-merge (generation confidence from token log-probabilities, base-model gradient cosines, and LoRA-delta geometry computed exactly in the adapters' low-rank factor space) and evaluates merges on held-out tests. Its rows are not independent, because parents share task-data seeds across conditions, so inference is condition-clustered and per-seed and leave-one-seed-out sensitivity are reported alongside; a committed script produces these statistics. Statistical, per-seed reproducibility is documented for the GPU tiers.

\textbf{Data availability.} All results artifacts (with content hashes) and the figures are in the project repository at \href{https://git.lab.gilest.ro/giorgio/MachineSex}{git.\allowbreak{}lab.\allowbreak{}gilest.\allowbreak{}ro/\allowbreak{}giorgio/\allowbreak{}MachineSex}, which will be archived under a DOI on publication; every figure in this paper regenerates from committed artifacts without re-simulation. The public datasets used are cited where they are introduced.

\textbf{Code availability.} All code, configs, seeds, and a one-command reproduction script are in the same repository.

\section*{Acknowledgements}

This work was done in close collaboration with Claude Opus 5 and Claude Fable 5.1 (Anthropic). I conceived the framework and the population-genetic reading, chose the questions and the experiments, set the predictions and falsifiers, directed every stage, judged the results and edited the text; the models contributed to the experimental design, wrote the code and ran the experiments under my direction, performed the analyses and wrote the first draft of the text. I take full responsibility for the content. I thank Imperial College London for funding, and its Research Computing Service for the high-performance computing resources (the CX3 cluster) on which the language-model experiments were run.

\section*{Author contributions}

G.F.G. conceived the framework, designed the study, directed and judged every experiment, and edited the manuscript. The contribution of the AI systems used is described in the Acknowledgements.

\section*{Competing interests}

The author declares no competing interests.

\section*{References}

\begin{enumerate}
\item B. Laufer, H. Oderinwale, J. Kleinberg, Anatomy of a machine learning ecosystem: 2 million models on Hugging Face. arXiv [Preprint] (2025). \href{https://doi.org/10.48550/arXiv.2508.06811}{https:/\allowbreak{}/\allowbreak{}doi.org/\allowbreak{}10.48550/\allowbreak{}arXiv.\allowbreak{}2508.06811}.
\item E. Horwitz, A. Shul, Y. Hoshen, Unsupervised model tree heritage recovery. \emph{Int. Conf. Learn. Represent.} (2025). \href{https://doi.org/10.48550/arXiv.2405.18432}{https:/\allowbreak{}/\allowbreak{}doi.org/\allowbreak{}10.48550/\allowbreak{}arXiv.\allowbreak{}2405.18432}.
\item W. Jiang, et al., PeaTMOSS: A dataset and initial analysis of pre-trained models in open-source software. \emph{Proc. Int. Conf. Min. Softw. Repos.} (2024). \href{https://doi.org/10.48550/arXiv.2402.00699}{https:/\allowbreak{}/\allowbreak{}doi.org/\allowbreak{}10.48550/\allowbreak{}arXiv.\allowbreak{}2402.00699}.
\item P. Yadav, D. Tam, L. Choshen, C. Raffel, M. Bansal, TIES-Merging: Resolving interference when merging models. \emph{Adv. Neural Inf. Process. Syst.} \textbf{36} (2023). \href{https://doi.org/10.48550/arXiv.2306.01708}{https:/\allowbreak{}/\allowbreak{}doi.org/\allowbreak{}10.48550/\allowbreak{}arXiv.\allowbreak{}2306.01708}.
\item T. Akiba, M. Shing, Y. Tang, Q. Sun, D. Ha, Evolutionary optimization of model merging recipes. \emph{Nat. Mach. Intell.} \textbf{7}, 195--204 (2025).
\item C. Goddard, et al., Arcee's MergeKit: A toolkit for merging large language models. \emph{Proc. Conf. Empir. Methods Nat. Lang. Process. (Industry Track)}, 477--485 (2024). \href{https://doi.org/10.48550/arXiv.2403.13257}{https:/\allowbreak{}/\allowbreak{}doi.org/\allowbreak{}10.48550/\allowbreak{}arXiv.\allowbreak{}2403.13257}.
\item E. Yang, et al., Model merging in LLMs, MLLMs, and beyond: Methods, theories, applications, and opportunities. \emph{ACM Comput. Surv.} \textbf{58}, 1--41 (2026). \href{https://doi.org/10.1145/\penalty5000{}3787849}{https:/\allowbreak{}/\allowbreak{}doi.org/\allowbreak{}10.1145/\allowbreak{}\penalty5000{}3787849}.
\item Y. Zhang, et al., Nature-inspired population-based evolution of large language models. \emph{Proc. Annu. Meet. Assoc. Comput. Linguist.}, 44187--44205 (2026). \href{https://doi.org/10.18653/v1/\penalty5000{}2026.acl-long.2044}{https:/\allowbreak{}/\allowbreak{}doi.org/\allowbreak{}10.18653/\allowbreak{}v1/\allowbreak{}\penalty5000{}2026.acl-long.2044}.
\item J. Abrantes, et al., Competition and attraction improve model fusion. \emph{Proc. Genet. Evol. Comput. Conf.} (2025). \href{https://doi.org/10.48550/arXiv.2508.16204}{https:/\allowbreak{}/\allowbreak{}doi.org/\allowbreak{}10.48550/\allowbreak{}arXiv.\allowbreak{}2508.16204}.
\item Y. Hu, Y. Yao, N. Zhang, H. Chen, S. Deng, Exploring model kinship for merging large language models. \emph{Findings Assoc. Comput. Linguist.: EMNLP}, 596--625 (2025). \href{https://doi.org/10.18653/v1/\penalty5000{}2025.findings-emnlp.32}{https:/\allowbreak{}/\allowbreak{}doi.org/\allowbreak{}10.18653/\allowbreak{}v1/\allowbreak{}\penalty5000{}2025.findings-emnlp.32}.
\item V. Subramaniam, et al., Multiagent finetuning: Self improvement with diverse reasoning chains. \emph{Int. Conf. Learn. Represent.} (2025). \href{https://doi.org/10.48550/arXiv.2501.05707}{https:/\allowbreak{}/\allowbreak{}doi.org/\allowbreak{}10.48550/\allowbreak{}arXiv.\allowbreak{}2501.05707}.
\item NVIDIA (B. Adler, et al.), Nemotron-4 340B technical report. arXiv [Preprint] (2024). \href{https://doi.org/10.48550/arXiv.2406.11704}{https:/\allowbreak{}/\allowbreak{}doi.org/\allowbreak{}10.48550/\allowbreak{}arXiv.\allowbreak{}2406.11704}.
\item M. Abdin, et al., Phi-4 technical report. arXiv [Preprint] (2024). \href{https://doi.org/10.48550/arXiv.2412.08905}{https:/\allowbreak{}/\allowbreak{}doi.org/\allowbreak{}10.48550/\allowbreak{}arXiv.\allowbreak{}2412.08905}.
\item Y. Wang, et al., Self-Instruct: Aligning language models with self-generated instructions. \emph{Proc. Annu. Meet. Assoc. Comput. Linguist.} (2023). \href{https://doi.org/10.48550/arXiv.2212.10560}{https:/\allowbreak{}/\allowbreak{}doi.org/\allowbreak{}10.48550/\allowbreak{}arXiv.\allowbreak{}2212.10560}.
\item B. Thompson, et al., A shocking amount of the web is machine translated: Insights from multi-way parallelism. \emph{Findings Assoc. Comput. Linguist.: ACL} (2024). \href{https://doi.org/10.48550/arXiv.2401.05749}{https:/\allowbreak{}/\allowbreak{}doi.org/\allowbreak{}10.48550/\allowbreak{}arXiv.\allowbreak{}2401.05749}.
\item W. Liang, et al., Monitoring AI-modified content at scale: A case study on the impact of ChatGPT on AI conference peer reviews. \emph{Proc. Int. Conf. Mach. Learn.} (2024). \href{https://doi.org/10.48550/arXiv.2403.07183}{https:/\allowbreak{}/\allowbreak{}doi.org/\allowbreak{}10.48550/\allowbreak{}arXiv.\allowbreak{}2403.07183}.
\item P. Villalobos, et al., Position: Will we run out of data? Limits of LLM scaling based on human-generated data. \emph{Proc. Int. Conf. Mach. Learn.} (2024). \href{https://doi.org/10.48550/arXiv.2211.04325}{https:/\allowbreak{}/\allowbreak{}doi.org/\allowbreak{}10.48550/\allowbreak{}arXiv.\allowbreak{}2211.04325}.
\item L. Brinkmann, et al., Machine culture. \emph{Nat. Hum. Behav.} \textbf{7}, 1855--1868 (2023).
\item J. S. Park, et al., Generative agents: Interactive simulacra of human behavior. \emph{Proc. ACM Symp. User Interface Softw. Technol.} (2023). \href{https://doi.org/10.48550/arXiv.2304.03442}{https:/\allowbreak{}/\allowbreak{}doi.org/\allowbreak{}10.48550/\allowbreak{}arXiv.\allowbreak{}2304.03442}.
\item T. Guo, et al., Large language model based multi-agents: A survey of progress and challenges. \emph{Proc. Int. Joint Conf. Artif. Intell.} (2024). \href{https://doi.org/10.48550/arXiv.2402.01680}{https:/\allowbreak{}/\allowbreak{}doi.org/\allowbreak{}10.48550/\allowbreak{}arXiv.\allowbreak{}2402.01680}.
\item N. Tomasev, et al., Virtual agent economies. arXiv [Preprint] (2025). \href{https://doi.org/10.48550/arXiv.2509.10147}{https:/\allowbreak{}/\allowbreak{}doi.org/\allowbreak{}10.48550/\allowbreak{}arXiv.\allowbreak{}2509.10147}.
\item A. Livnat, C. Papadimitriou, Sex as an algorithm: The theory of evolution under the lens of computation. \emph{Commun. ACM} \textbf{59}, 84--93 (2016).
\item I. Shumailov, et al., AI models collapse when trained on recursively generated data. \emph{Nature} \textbf{631}, 755--759 (2024).
\item J. P. Crutchfield, S. Whalen, Structural drift: The population dynamics of sequential learning. \emph{PLOS Comput. Biol.} \textbf{8}, e1002510 (2012).
\item S. Riis, Drift and selection in LLM text ecosystems. arXiv [Preprint] (2026). \href{https://doi.org/10.48550/arXiv.2604.08554}{https:/\allowbreak{}/\allowbreak{}doi.org/\allowbreak{}10.48550/\allowbreak{}arXiv.\allowbreak{}2604.08554}.
\item M. Benati, A. Londei, D. Lanzieri, V. Loreto, First-extinction law for resampling processes. arXiv [Preprint] (2025). \href{https://doi.org/10.48550/arXiv.2509.20101}{https:/\allowbreak{}/\allowbreak{}doi.org/\allowbreak{}10.48550/\allowbreak{}arXiv.\allowbreak{}2509.20101}.
\item Y. Yoon, D. Hu, I. Weissburg, Y. Qin, H. Jeong, Model collapse in the self-consuming chain of diffusion finetuning: A novel perspective from quantitative trait modeling. \emph{ICLR Workshop on Navigating and Addressing Data Problems for Foundation Models} (2025). \href{https://doi.org/10.48550/arXiv.2407.17493}{https:/\allowbreak{}/\allowbreak{}doi.org/\allowbreak{}10.48550/\allowbreak{}arXiv.\allowbreak{}2407.17493}.
\item M. McCloskey, N. J. Cohen, Catastrophic interference in connectionist networks: The sequential learning problem. \emph{Psychol. Learn. Motiv.} \textbf{24}, 109--165 (1989).
\item R. M. French, Catastrophic forgetting in connectionist networks. \emph{Trends Cogn. Sci.} \textbf{3}, 128--135 (1999).
\item H. J. Muller, The relation of recombination to mutational advance. \emph{Mutat. Res.} \textbf{1}, 2--9 (1964).
\item S. Alemohammad, et al., Self-consuming generative models go MAD. \emph{Int. Conf. Learn. Represent.} (2024). \href{https://doi.org/10.48550/arXiv.2307.01850}{https:/\allowbreak{}/\allowbreak{}doi.org/\allowbreak{}10.48550/\allowbreak{}arXiv.\allowbreak{}2307.01850}.
\item Q. Bertrand, A. J. Bose, A. Duplessis, M. Jiralerspong, G. Gidel, On the stability of iterative retraining of generative models on their own data. \emph{Int. Conf. Learn. Represent.} (2024). \href{https://doi.org/10.48550/arXiv.2310.00429}{https:/\allowbreak{}/\allowbreak{}doi.org/\allowbreak{}10.48550/\allowbreak{}arXiv.\allowbreak{}2310.00429}.
\item B. Yi, Q. Liu, Y. Cheng, H. Xu, Escaping model collapse via synthetic data verification. arXiv [Preprint] (2025). \href{https://doi.org/10.48550/arXiv.2510.16657}{https:/\allowbreak{}/\allowbreak{}doi.org/\allowbreak{}10.48550/\allowbreak{}arXiv.\allowbreak{}2510.16657}.
\item M. Gerstgrasser, et al., Is model collapse inevitable? Breaking the curse of recursion by accumulating real and synthetic data. \emph{Conf. Lang. Model.} (2024). \href{https://doi.org/10.48550/arXiv.2404.01413}{https:/\allowbreak{}/\allowbreak{}doi.org/\allowbreak{}10.48550/\allowbreak{}arXiv.\allowbreak{}2404.01413}.
\item L. S. Mills, F. W. Allendorf, The one-migrant-per-generation rule in conservation and management. \emph{Conserv. Biol.} \textbf{10}, 1509--1518 (1996).
\item M. Wortsman, et al., Model soups: Averaging weights of multiple fine-tuned models improves accuracy without increasing inference time. \emph{Proc. Int. Conf. Mach. Learn.} (2022). \href{https://doi.org/10.48550/arXiv.2203.05482}{https:/\allowbreak{}/\allowbreak{}doi.org/\allowbreak{}10.48550/\allowbreak{}arXiv.\allowbreak{}2203.05482}.
\item R. A. Fisher, \emph{The Genetical Theory of Natural Selection} (Clarendon Press, 1930).
\item H. J. Muller, Some genetic aspects of sex. \emph{Am. Nat.} \textbf{66}, 118--138 (1932).
\item S. P. Otto, M. W. Feldman, Deleterious mutations, variable epistatic interactions, and the evolution of recombination. \emph{Theor. Popul. Biol.} \textbf{51}, 134--147 (1997).
\item A. R. Templeton, ``Coadaptation and outbreeding depression'' in \emph{Conservation Biology: The Science of Scarcity and Diversity}, M. E. Soulé, Ed. (Sinauer, 1986), pp. 105--116.
\item M. Tomassini, \emph{Spatially Structured Evolutionary Algorithms: Artificial Evolution in Space and Time} (Springer, 2005).
\item H. A. Orr, The population genetics of speciation: The evolution of hybrid incompatibilities. \emph{Genetics} \textbf{139}, 1805--1813 (1995).
\item H. A. Orr, M. Turelli, The evolution of postzygotic isolation: Accumulating Dobzhansky--Muller incompatibilities. \emph{Evolution} \textbf{55}, 1085--1094 (2001).
\item S. K. Ainsworth, J. Hayase, S. Srinivasa, Git Re-Basin: Merging models modulo permutation symmetries. \emph{Int. Conf. Learn. Represent.} (2023). \href{https://doi.org/10.48550/arXiv.2209.04836}{https:/\allowbreak{}/\allowbreak{}doi.org/\allowbreak{}10.48550/\allowbreak{}arXiv.\allowbreak{}2209.04836}.
\item E. Sharma, D. M. Roy, G. K. Dziugaite, The non-local model merging problem: Permutation symmetries and variance collapse. arXiv [Preprint] (2024). \href{https://doi.org/10.48550/arXiv.2410.12766}{https:/\allowbreak{}/\allowbreak{}doi.org/\allowbreak{}10.48550/\allowbreak{}arXiv.\allowbreak{}2410.12766}.
\item K. Jordan, H. Sedghi, O. Saukh, R. Entezari, B. Neyshabur, REPAIR: REnormalizing permuted activations for interpolation repair. \emph{Int. Conf. Learn. Represent.} (2023). \href{https://doi.org/10.48550/arXiv.2211.08403}{https:/\allowbreak{}/\allowbreak{}doi.org/\allowbreak{}10.48550/\allowbreak{}arXiv.\allowbreak{}2211.08403}.
\item G. Stoica, et al., ZipIt! Merging models from different tasks without training. \emph{Int. Conf. Learn. Represent.} (2024). \href{https://doi.org/10.48550/arXiv.2305.03053}{https:/\allowbreak{}/\allowbreak{}doi.org/\allowbreak{}10.48550/\allowbreak{}arXiv.\allowbreak{}2305.03053}.
\item A. Kleiman, G. K. Dziugaite, J. Frankle, S. Kakade, M. Paul, Soup to go: Mitigating forgetting during continual learning with model averaging. arXiv [Preprint] (2025). \href{https://doi.org/10.48550/arXiv.2501.05559}{https:/\allowbreak{}/\allowbreak{}doi.org/\allowbreak{}10.48550/\allowbreak{}arXiv.\allowbreak{}2501.05559}.
\item X. Yuan, et al., Superficial self-improved reasoners benefit from model merging. \emph{Proc. Conf. Empir. Methods Nat. Lang. Process.}, 5912--5932 (2025). \href{https://doi.org/10.18653/v1/\penalty5000{}2025.emnlp-main.301}{https:/\allowbreak{}/\allowbreak{}doi.org/\allowbreak{}10.18653/\allowbreak{}v1/\allowbreak{}\penalty5000{}2025.emnlp-main.301}.
\item N. H. Barton, A general model for the evolution of recombination. \emph{Genet. Res.} \textbf{65}, 123--144 (1995).
\item S. P. Otto, T. Lenormand, Resolving the paradox of sex and recombination. \emph{Nat. Rev. Genet.} \textbf{3}, 252--261 (2002).
\item L. Altenberg, M. W. Feldman, Selection, generalized transmission and the evolution of modifier genes. I. The reduction principle. \emph{Genetics} \textbf{117}, 559--572 (1987).
\item T. Fukuda, H. Kera, K. Kawamoto, Adapter merging with centroid prototype mapping for scalable class-incremental learning. \emph{Proc. IEEE/CVF Conf. Comput. Vis. Pattern Recognit.} (2025). \href{https://doi.org/10.48550/arXiv.2412.18219}{https:/\allowbreak{}/\allowbreak{}doi.org/\allowbreak{}10.48550/\allowbreak{}arXiv.\allowbreak{}2412.18219}.
\item D. Shenaj, O. Bohdal, T. Ceritli, M. Ozay, P. Zanuttigh, U. Michieli, K-Merge: Online continual merging of adapters for on-device large language models. \emph{Proc. Annu. Meet. Assoc. Comput. Linguist.}, 3013--3029 (2026). \href{https://doi.org/10.18653/v1/\penalty5000{}2026.acl-long.137}{https:/\allowbreak{}/\allowbreak{}doi.org/\allowbreak{}10.18653/\allowbreak{}v1/\allowbreak{}\penalty5000{}2026.acl-long.137}.
\item J. Lehman, K. O. Stanley, Abandoning objectives: Evolution through the search for novelty alone. \emph{Evol. Comput.} \textbf{19}, 189--223 (2011).
\item S. Wright, Evolution in Mendelian populations. \emph{Genetics} \textbf{16}, 97--159 (1931).
\item E. Dohmatob, Y. Feng, P. Yang, F. Charton, J. Kempe, A tale of tails: Model collapse as a change of scaling laws. \emph{Proc. Int. Conf. Mach. Learn.} (2024). \href{https://doi.org/10.48550/arXiv.2402.07043}{https:/\allowbreak{}/\allowbreak{}doi.org/\allowbreak{}10.48550/\allowbreak{}arXiv.\allowbreak{}2402.07043}.
\item E. Dohmatob, Y. Feng, A. Subramonian, J. Kempe, Strong model collapse. \emph{Int. Conf. Learn. Represent.} (2025). \href{https://doi.org/10.48550/arXiv.2410.04840}{https:/\allowbreak{}/\allowbreak{}doi.org/\allowbreak{}10.48550/\allowbreak{}arXiv.\allowbreak{}2410.04840}.
\item A. Garg, S. Bhattacharya, P. Sur, Preventing model collapse under overparametrization: Optimal mixing ratios for interpolation learning and ridge regression. arXiv [Preprint] (2025). \href{https://doi.org/10.48550/arXiv.2509.22341}{https:/\allowbreak{}/\allowbreak{}doi.org/\allowbreak{}10.48550/\allowbreak{}arXiv.\allowbreak{}2509.22341}.
\item A. T. Suresh, A. Thangaraj, A. N. K. Khandavally, Rate of model collapse in recursive training. \emph{Proc. Int. Conf. Artif. Intell. Stat.}, PMLR \textbf{258}, 1396--1404 (2025). \href{https://doi.org/10.48550/arXiv.2412.17646}{https:/\allowbreak{}/\allowbreak{}doi.org/\allowbreak{}10.48550/\allowbreak{}arXiv.\allowbreak{}2412.17646}.
\item J. Kazdan, et al., Collapse or thrive? Perils and promises of synthetic data in a self-generating world. \emph{Proc. Int. Conf. Mach. Learn.}, PMLR \textbf{267}, 29469--29494 (2025). \href{https://doi.org/10.48550/arXiv.2410.16713}{https:/\allowbreak{}/\allowbreak{}doi.org/\allowbreak{}10.48550/\allowbreak{}arXiv.\allowbreak{}2410.16713}.
\item A. Souly, et al., Poisoning attacks on LLMs require a near-constant number of poison samples. arXiv [Preprint] (2025). \href{https://doi.org/10.48550/arXiv.2510.07192}{https:/\allowbreak{}/\allowbreak{}doi.org/\allowbreak{}10.48550/\allowbreak{}arXiv.\allowbreak{}2510.07192}.
\item F. Jenkin, The origin of species [review]. \emph{North Br. Rev.} \textbf{46}, 277--318 (1867).
\item M. Bulmer, Did Jenkin's swamping argument invalidate Darwin's theory of natural selection? \emph{Br. J. Hist. Sci.} \textbf{37}, 281--297 (2004).
\item A. Malinin, B. Mlodozeniec, M. Gales, Ensemble distribution distillation. \emph{Int. Conf. Learn. Represent.} (2020). \href{https://doi.org/10.48550/arXiv.1905.00076}{https:/\allowbreak{}/\allowbreak{}doi.org/\allowbreak{}10.48550/\allowbreak{}arXiv.\allowbreak{}1905.00076}.
\item M. Li, et al., Branch-Train-Merge: Embarrassingly parallel training of expert language models. arXiv [Preprint] (2022). \href{https://doi.org/10.48550/arXiv.2208.03306}{https:/\allowbreak{}/\allowbreak{}doi.org/\allowbreak{}10.48550/\allowbreak{}arXiv.\allowbreak{}2208.03306}.
\item X. Yuan, et al., Behavior knowledge merge in reinforced agentic models. \emph{Proc. Annu. Meet. Assoc. Comput. Linguist.}, 33007--33028 (2026). \href{https://doi.org/10.18653/v1/\penalty5000{}2026.acl-long.1524}{https:/\allowbreak{}/\allowbreak{}doi.org/\allowbreak{}10.18653/\allowbreak{}v1/\allowbreak{}\penalty5000{}2026.acl-long.1524}.
\item J. Pari, S. Jelassi, P. Agrawal, Collective model intelligence requires compatible specialization. arXiv [Preprint] (2024). \href{https://doi.org/10.48550/arXiv.2411.02207}{https:/\allowbreak{}/\allowbreak{}doi.org/\allowbreak{}10.48550/\allowbreak{}arXiv.\allowbreak{}2411.02207}.
\item E. J. Hu, et al., LoRA: Low-rank adaptation of large language models. \emph{Int. Conf. Learn. Represent.} (2022). \href{https://doi.org/10.48550/arXiv.2106.09685}{https:/\allowbreak{}/\allowbreak{}doi.org/\allowbreak{}10.48550/\allowbreak{}arXiv.\allowbreak{}2106.09685}.
\item L. Yu, B. Yu, H. Yu, F. Huang, Y. Li, Language models are super Mario: Absorbing abilities from homologous models as a free lunch. \emph{Proc. Int. Conf. Mach. Learn.} (2024). \href{https://doi.org/10.48550/arXiv.2311.03099}{https:/\allowbreak{}/\allowbreak{}doi.org/\allowbreak{}10.48550/\allowbreak{}arXiv.\allowbreak{}2311.03099}.
\item S. A. Kauffman, S. Levin, Towards a general theory of adaptive walks on rugged landscapes. \emph{J. Theor. Biol.} \textbf{128}, 11--45 (1987).
\item D. Marczak, B. Twardowski, T. Trzciński, S. Cygert, MagMax: Leveraging model merging for seamless continual learning. \emph{Proc. Eur. Conf. Comput. Vis.} (2024). \href{https://doi.org/10.48550/arXiv.2407.06322}{https:/\allowbreak{}/\allowbreak{}doi.org/\allowbreak{}10.48550/\allowbreak{}arXiv.\allowbreak{}2407.06322}.
\item S. Dziadzio, et al., How to merge your multimodal models over time? \emph{Proc. IEEE/CVF Conf. Comput. Vis. Pattern Recognit.} (2025). \href{https://doi.org/10.48550/arXiv.2412.06712}{https:/\allowbreak{}/\allowbreak{}doi.org/\allowbreak{}10.48550/\allowbreak{}arXiv.\allowbreak{}2412.06712}.
\item A. Williams, N. Nangia, S. R. Bowman, A broad-coverage challenge corpus for sentence understanding through inference. \emph{Proc. Conf. North Am. Chapter Assoc. Comput. Linguist. Hum. Lang. Technol.}, 1112--1122 (2018).
\item P. Clark, et al., Think you have solved question answering? Try ARC, the AI2 Reasoning Challenge. arXiv [Preprint] (2018). \href{https://doi.org/10.48550/arXiv.1803.05457}{https:/\allowbreak{}/\allowbreak{}doi.org/\allowbreak{}10.48550/\allowbreak{}arXiv.\allowbreak{}1803.05457}.
\item R. Zellers, A. Holtzman, Y. Bisk, A. Farhadi, Y. Choi, HellaSwag: Can a machine really finish your sentence? \emph{Proc. Annu. Meet. Assoc. Comput. Linguist.}, 4791--4800 (2019).
\item P. Rajpurkar, J. Zhang, K. Lopyrev, P. Liang, SQuAD: 100,000+ questions for machine comprehension of text. \emph{Proc. Conf. Empir. Methods Nat. Lang. Process.}, 2383--2392 (2016).
\item C. Clark, et al., BoolQ: Exploring the surprising difficulty of natural yes/no questions. \emph{Proc. Conf. North Am. Chapter Assoc. Comput. Linguist. Hum. Lang. Technol.}, 2924--2936 (2019).
\item K. Sakaguchi, R. Le Bras, C. Bhagavatula, Y. Choi, WinoGrande: An adversarial Winograd schema challenge at scale. \emph{Proc. AAAI Conf. Artif. Intell.} \textbf{34}, 8732--8740 (2020).
\item L. Thede, K. Roth, M. Bethge, Z. Akata, T. Hartvigsen, WikiBigEdit: Understanding the limits of lifelong knowledge editing in LLMs. \emph{Proc. Int. Conf. Mach. Learn.} (2025). \href{https://doi.org/10.48550/arXiv.2503.05683}{https:/\allowbreak{}/\allowbreak{}doi.org/\allowbreak{}10.48550/\allowbreak{}arXiv.\allowbreak{}2503.05683}.
\item S. Clemente, et al., In praise of stubbornness: An empirical case for cognitive-dissonance aware continual update of knowledge in LLMs. arXiv [Preprint] (2025). \href{https://doi.org/10.48550/arXiv.2502.04390}{https:/\allowbreak{}/\allowbreak{}doi.org/\allowbreak{}10.48550/\allowbreak{}arXiv.\allowbreak{}2502.04390}.
\item J. Störk, Interference and retention in continual learning. arXiv [Preprint] (2026). \href{https://doi.org/10.48550/arXiv.2607.09202}{https:/\allowbreak{}/\allowbreak{}doi.org/\allowbreak{}10.48550/\allowbreak{}arXiv.\allowbreak{}2607.09202}.
\item T. Li, Z. Shen, Scaling linear mode connectivity and merging to billion-parameter pretrained transformers. arXiv [Preprint] (2026). \href{https://doi.org/10.48550/arXiv.2606.23607}{https:/\allowbreak{}/\allowbreak{}doi.org/\allowbreak{}10.48550/\allowbreak{}arXiv.\allowbreak{}2606.23607}.
\item N. Kozodoi, Z. Afolabi, J. Butler, Are we merging the right models? Impact of expert training duration on model merging for LLMs. arXiv [Preprint] (2026). \href{https://doi.org/10.48550/arXiv.2607.11997}{https:/\allowbreak{}/\allowbreak{}doi.org/\allowbreak{}10.48550/\allowbreak{}arXiv.\allowbreak{}2607.11997}.
\item S. Horoi, G. Wolf, E. Belilovsky, G. K. Dziugaite, From memorization to parameter interference: How overtraining experts harms model merging. \emph{Proc. Int. Conf. Mach. Learn.} (2026). \href{https://doi.org/10.48550/arXiv.2506.14126}{https:/\allowbreak{}/\allowbreak{}doi.org/\allowbreak{}10.48550/\allowbreak{}arXiv.\allowbreak{}2506.14126}.
\item L. Zhou, B. Zhao, R. Yu, E. Rodolà, Demystifying mergeability: Interpretable properties to predict model merging success. arXiv [Preprint] (2026). \href{https://doi.org/10.48550/arXiv.2601.22285}{https:/\allowbreak{}/\allowbreak{}doi.org/\allowbreak{}10.48550/\allowbreak{}arXiv.\allowbreak{}2601.22285}.
\item Y. Cao, et al., An empirical study and theoretical explanation on task-level model-merging collapse. arXiv [Preprint] (2026). \href{https://doi.org/10.48550/arXiv.2603.09463}{https:/\allowbreak{}/\allowbreak{}doi.org/\allowbreak{}10.48550/\allowbreak{}arXiv.\allowbreak{}2603.09463}.
\item C. Zhu, X. Li, T. Cai, When do task vectors interfere? Mapping the validity boundaries of weight-space composition. arXiv [Preprint] (2026). \href{https://doi.org/10.48550/arXiv.2608.09490}{https:/\allowbreak{}/\allowbreak{}doi.org/\allowbreak{}10.48550/\allowbreak{}arXiv.\allowbreak{}2608.09490}.
\item T. Scialom, T. Chakrabarty, S. Muresan, Fine-tuned language models are continual learners. \emph{Proc. Conf. Empir. Methods Nat. Lang. Process.} (2022). \href{https://doi.org/10.48550/arXiv.2205.12393}{https:/\allowbreak{}/\allowbreak{}doi.org/\allowbreak{}10.48550/\allowbreak{}arXiv.\allowbreak{}2205.12393}.
\item A. Ibrahim, et al., Simple and scalable strategies to continually pre-train large language models. \emph{Trans. Mach. Learn. Res.} (2024). \href{https://doi.org/10.48550/arXiv.2403.08763}{https:/\allowbreak{}/\allowbreak{}doi.org/\allowbreak{}10.48550/\allowbreak{}arXiv.\allowbreak{}2403.08763}.
\item A. Robins, Catastrophic forgetting, rehearsal and pseudorehearsal. \emph{Connect. Sci.} \textbf{7}, 123--146 (1995).
\item H. Shin, J. K. Lee, J. Kim, J. Kim, Continual learning with deep generative replay. \emph{Adv. Neural Inf. Process. Syst.} \textbf{30} (2017). \href{https://doi.org/10.48550/arXiv.1705.08690}{https:/\allowbreak{}/\allowbreak{}doi.org/\allowbreak{}10.48550/\allowbreak{}arXiv.\allowbreak{}1705.08690}.
\item Y. Feng, et al., Beyond model collapse: Scaling up with synthesized data requires verification. \emph{Int. Conf. Learn. Represent.} (2025). \href{https://doi.org/10.48550/arXiv.2406.07515}{https:/\allowbreak{}/\allowbreak{}doi.org/\allowbreak{}10.48550/\allowbreak{}arXiv.\allowbreak{}2406.07515}.
\item A. A. Rusu, et al., Progressive neural networks. arXiv [Preprint] (2016). \href{https://doi.org/10.48550/arXiv.1606.04671}{https:/\allowbreak{}/\allowbreak{}doi.org/\allowbreak{}10.48550/\allowbreak{}arXiv.\allowbreak{}1606.04671}.
\item D. Biderman, et al., LoRA learns less and forgets less. \emph{Trans. Mach. Learn. Res.} (2024). \href{https://doi.org/10.48550/arXiv.2405.09673}{https:/\allowbreak{}/\allowbreak{}doi.org/\allowbreak{}10.48550/\allowbreak{}arXiv.\allowbreak{}2405.09673}.
\item J. L. McClelland, B. L. McNaughton, R. C. O'Reilly, Why there are complementary learning systems in the hippocampus and neocortex: Insights from the successes and failures of connectionist models of learning and memory. \emph{Psychol. Rev.} \textbf{102}, 419--457 (1995).
\item D. Kumaran, D. Hassabis, J. L. McClelland, What learning systems do intelligent agents need? Complementary learning systems theory updated. \emph{Trends Cogn. Sci.} \textbf{20}, 512--534 (2016).
\item J. Schwarz, et al., Progress \& Compress: A scalable framework for continual learning. \emph{Proc. Int. Conf. Mach. Learn.} (2018).
\item G. Ilharco, et al., Editing models with task arithmetic. \emph{Int. Conf. Learn. Represent.} (2023). \href{https://doi.org/10.48550/arXiv.2212.04089}{https:/\allowbreak{}/\allowbreak{}doi.org/\allowbreak{}10.48550/\allowbreak{}arXiv.\allowbreak{}2212.04089}.
\item A. Alexandrov, et al., Mitigating catastrophic forgetting in language transfer via model merging. \emph{Findings Assoc. Comput. Linguist.: EMNLP} (2024). \href{https://doi.org/10.48550/arXiv.2407.08699}{https:/\allowbreak{}/\allowbreak{}doi.org/\allowbreak{}10.48550/\allowbreak{}arXiv.\allowbreak{}2407.08699}.
\item M. Toneva, et al., An empirical study of example forgetting during deep neural network learning. \emph{Int. Conf. Learn. Represent.} (2019). \href{https://doi.org/10.48550/arXiv.1812.05159}{https:/\allowbreak{}/\allowbreak{}doi.org/\allowbreak{}10.48550/\allowbreak{}arXiv.\allowbreak{}1812.05159}.
\item N. Kandpal, H. Deng, A. Roberts, E. Wallace, C. Raffel, Large language models struggle to learn long-tail knowledge. \emph{Proc. Int. Conf. Mach. Learn.} (2023). \href{https://doi.org/10.48550/arXiv.2211.08411}{https:/\allowbreak{}/\allowbreak{}doi.org/\allowbreak{}10.48550/\allowbreak{}arXiv.\allowbreak{}2211.08411}.
\item X. Liu, et al., Long-tailed class incremental learning. \emph{Proc. Eur. Conf. Comput. Vis.} (2022). \href{https://doi.org/10.48550/arXiv.2210.00266}{https:/\allowbreak{}/\allowbreak{}doi.org/\allowbreak{}10.48550/\allowbreak{}arXiv.\allowbreak{}2210.00266}.
\item S. Kotha, J. M. Springer, A. Raghunathan, Understanding catastrophic forgetting in language models via implicit inference. \emph{Int. Conf. Learn. Represent.} (2024). \href{https://doi.org/10.48550/arXiv.2309.10105}{https:/\allowbreak{}/\allowbreak{}doi.org/\allowbreak{}10.48550/\allowbreak{}arXiv.\allowbreak{}2309.10105}.
\item Y. Guo, G. Shang, M. Vazirgiannis, C. Clavel, The curious decline of linguistic diversity: Training language models on synthetic text. \emph{Findings Assoc. Comput. Linguist.: NAACL} (2024). \href{https://doi.org/10.48550/arXiv.2311.09807}{https:/\allowbreak{}/\allowbreak{}doi.org/\allowbreak{}10.48550/\allowbreak{}arXiv.\allowbreak{}2311.09807}.
\item V. Padmakumar, H. He, Does writing with language models reduce content diversity? \emph{Int. Conf. Learn. Represent.} (2024). \href{https://doi.org/10.48550/arXiv.2309.05196}{https:/\allowbreak{}/\allowbreak{}doi.org/\allowbreak{}10.48550/\allowbreak{}arXiv.\allowbreak{}2309.05196}.
\item A. R. Doshi, O. P. Hauser, Generative AI enhances individual creativity but reduces the collective diversity of novel content. \emph{Sci. Adv.} \textbf{10}, eadn5290 (2024).
\end{enumerate}

\end{document}